\documentclass{article}

\usepackage{microtype}
\usepackage{graphicx}
\usepackage{subfigure}
\usepackage{booktabs} % for professional tables
\usepackage{multirow} 
\usepackage[table]{xcolor}
\usepackage{paralist}   % 更紧凑的列表
\usepackage{empheq}

\usepackage{hyperref}

\usepackage[accepted]{icml2025}

\usepackage{amsmath}
\usepackage{amssymb}
\usepackage{mathtools}
\usepackage{amsthm}

\usepackage[capitalize,noabbrev]{cleveref}

\theoremstyle{plain}
\newtheorem{theorem}{Theorem}[section]
\newtheorem{proposition}[theorem]{Proposition}

\newtheorem{corollary}[theorem]{Corollary}
\theoremstyle{definition}
\newtheorem{definition}[theorem]{Definition}
\newtheorem{assumption}[theorem]{Assumption}
\theoremstyle{remark}

\crefname{section}{Sec.}{Secs.}
\Crefname{section}{Section}{Sections}
\crefname{table}{Tab.}{Tabs.}
\Crefname{table}{Table}{Tables}
\crefname{figure}{Fig.}{Figs.}
\Crefname{figure}{Figure}{Figures}
\crefname{equation}{Eq.}{Eqs.}
\Crefname{equation}{Equation}{Equations}
\crefname{assumption}{Assumption}{Assumptions}
\Crefname{assumption}{Assumption}{Assumptions}

\usepackage[textsize=tiny]{todonotes}

\icmltitlerunning{Dequantified Diffusion-Schr{\"o}dinger Bridge for Density Ratio Estimation}

\begin{document}

\twocolumn[
\icmltitle{Dequantified Diffusion-Schr{\"o}dinger Bridge for Density Ratio Estimation}





\begin{icmlauthorlist}
\icmlauthor{Wei Chen}{yyy}
\icmlauthor{Shigui Li}{yyy}
\icmlauthor{Jiacheng Li}{sch}
\icmlauthor{Junmei Yang}{sch}
\icmlauthor{John Paisley}{columbia}
\icmlauthor{Delu Zeng}{sch}

\end{icmlauthorlist}

\icmlaffiliation{yyy}{The School of Mathematics, South China University of Technology, Guangzhou 510006, China}
\icmlaffiliation{sch}{The School of Electronic and Information Engineering, South China University of Technology, Guangzhou 510006, China}
\icmlaffiliation{columbia}{The Department of Electrical Engineering, Columbia
University, New York, NY 10027, USA}  % jpaisley@columbia.edu

\icmlcorrespondingauthor{Delu Zeng}{dlzeng@scut.edu.cn}

\icmlkeywords{Machine Learning, ICML}

\vskip 0.3in
]



\printAffiliationsAndNotice{}  % leave blank if no need to mention equal contribution

\begin{abstract}
Density ratio estimation is fundamental to tasks involving $f$-divergences, yet existing methods often fail under significantly different distributions or inadequately overlapping supports --- the \textit{density-chasm} and the \textit{support-chasm} problems.
Additionally, prior approaches yield divergent time scores near boundaries, leading to instability.
We design $\textbf{D}^3\textbf{RE}$, a unified framework for \textbf{robust}, \textbf{stable} and \textbf{efficient} density ratio estimation.
We propose the \textit{dequantified diffusion bridge interpolant} (\textbf{DDBI}), which expands support coverage and stabilizes time scores via diffusion bridges and Gaussian dequantization.
Building on DDBI, the proposed \textit{dequantified Schr{\"o}dinger bridge interpolant} (\textbf{DSBI}) incorporates optimal transport to solve the Schr{\"o}dinger bridge problem, enhancing accuracy and efficiency.
Our method offers uniform approximation and bounded time scores in theory, and outperforms baselines empirically in mutual information and density estimation tasks.
Code is available at \href{https://github.com/Hoemr/Dequantified-Diffusion-Bridge-Density-Ratio-Estimation.git}{https://github.com/Hoemr/Dequantified-Diffusion-Bridge-Density-Ratio-Estimation.git}.

\end{abstract}

\section{Introduction}
Quantifying distributional discrepancies via $ f $-divergences-defined through density ratios $ r(\mathbf{x}) = q_1(\mathbf{x})/q_0(\mathbf{x})$ is foundational in tasks such as domain adaptation, generative modeling, and hypothesis testing. However, directly estimating $ r(\mathbf{x}) $ by modeling $ q_0 $ and $ q_1 $ becomes intractable in high dimensions, motivating density ratio estimation (DRE) methods that bypass explicit density modeling \citep{sugiyama2012density}. While DRE underpins modern techniques like mutual information estimation \citep{colombo2021novel} and likelihood-free inference \citep{thomas2022likelihood}, it struggles with a critical challenge known as the \textit{density-chasm } problem, where multi-modal or divergent distributions lead to unstable ratio estimates \citep{rhodes2020telescoping}.

Existing methods like telescoping ratio estimation (TRE) \citep{rhodes2020telescoping} and its continuous extension DRE-$ \infty $ \citep{choi2022density} estimate density ratios via intermediate steps. TRE improves accuracy by adding more intermediate variables, but increases model complexity linearly. DRE-$ \infty $ uses continuous-time score matching to avoid this, yet both face a core challenge in the \textit{support-chasm problem} (see \cref{definition:support-chasm-problem}), where $\mathsf{supp}(q_0) \cap \mathsf{supp}(q_1)$ is small or empty. This leads to inadequately overlapping supports and ill-defined ratios \citep{srivastava2023estimating}. 

To address the support-chasm problem, we unify the interpolation strategies in prior works as \textit{deterministic interpolants} (\textbf{DI}) and propose the \textit{diffusion bridge interpolant} (\textbf{DBI}), which uses diffusion bridges to enable diverse trajectory exploration and smooth transitions between distributions. 
By \cref{theorem:support-set-expansion} and  \cref{corollary:path-set-expansion}, DBI expands support coverage and trajectory sets beyond existing approaches.%, reducing bias in DRE \cref{corollary:reduced-estimation-error}.

A second challenge arises in prior methods: As $t \to 1^-$, the absolute time score $\mathbb{E}_{q_t}\left[|\partial_t \log q_t|\right]$ diverges for both DI and DBI (\cref{theorem:divergent-lower-bound-time-score}), leading to unstable estimations at the boundary.
To mitigate this, we propose \textit{Gaussian dequantization} (\textbf{GD}), which addresses boundary densities $q_0$ and $q_1$ via Gaussian convolution, ensuring $\mathbb{E}_{q_t}\left[|\partial_t \log q_t|\right]$ remains bounded over $t \in [0,1]$ (\cref{corollary:upper-bound-time-score-DDBI-DSBI}). The resulting \textit{dequantified diffusion bridge interpolant} (\textbf{DDBI}) balances robustness and computational efficiency.

To further reduce estimation error and improve efficiency, the \textit{dequantified Schr{\"o}dinger bridge interpolant} (\textbf{DSBI}) is derived by integrating DDBI with \textit{optimal transport rearrangement} (\textbf{OTR}), solving the Schr{\"o}dinger bridge problem  (\cref{proposition:solution-to-SB-problem}). 
Together, applying DDBI and DSBI to DRE leads to the \textit{\textbf{D}equantified \textbf{D}iffusion \textbf{B}ridge \textbf{D}ensity \textbf{R}atio \textbf{E}stimation} ($ \textbf{D}^3\textbf{RE} $) framework.  We summarize these developments in Table \ref{tab:my_label}.
% Dequantified Diffusion-Bridge Density Ratio Estimation

\begin{figure}[t]
    \centering
    \subfigure[DI (previous)]{\includegraphics[width=0.495\linewidth]{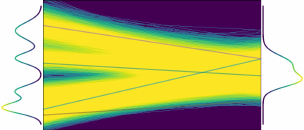}\label{fig:DI-concate}}
    \hfill
    \subfigure[DBI (ours)]{\includegraphics[width=0.495\linewidth]{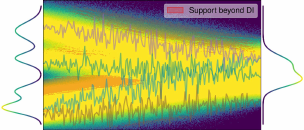}\label{fig:DBI-concate}}
    
    \subfigure[DDBI (ours) ]{\includegraphics[width=0.495\linewidth]{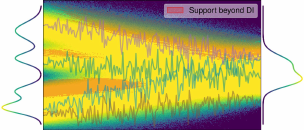}\label{fig:DDBI-concate}}
    \hfill
    \subfigure[DSBI (ours)]{\includegraphics[width=0.495\linewidth]{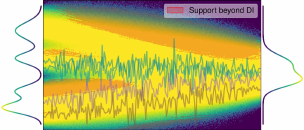}\label{fig:DSBI-concate}}
    
    \caption{Trajectory sets comparison among  DI, DBI, DDBI and DSBI. 
    Our methods yield \textit{\textbf{broader trajectory sets}}, with intermediate distributions exhibiting wider support than those of DI.
    The entropically regularized transport losses for subfigures (a-d) are $44.17$, $31.14$, $31.15$, and $8.26$, respectively (See \cref{eq:entropic-OT-problem} for details.).  Lower loss indicates increased path diversity.
    }
    \label{fig:compare-trajectory}
\end{figure}

Experimental results show that $\text{D}^3\text{RE}$ improves robustness and efficiency in downstream tasks such as density ratio estimation, mutual information estimation, and density estimation. 
\cref{fig:compare-trajectory} illustrates a comparison of interpolation strategies among DI, DBI, DDBI, and DSBI, with light blue points representing intermediate samples drawn from $q_0$ and $q_1$. Our proposed methods (DBI, DDBI, and DSBI) enable broader exploration of alternative trajectories, producing intermediate distributions with larger support compared to DI, consistent with \cref{theorem:support-set-expansion} and \cref{corollary:path-set-expansion} below.

\begin{table}[t]
    \centering
    \caption{Comparison of advantages of interpolants in this work. 
    % The second row indicates the performance benefits contributed by each module. % Both DDBI and DSBI are specific instances of our proposed D$^3$RE framework.
    }
     \vskip 0.1in
    \resizebox{\linewidth}{!}{
    \begin{tabular}{r|ccc}
    \hline
         & \textbf{Diffusion bridge} & \textbf{GD} & \textbf{OTR} \\
         & (robust \& stable) & (stable) & (efficient) \\
         \hline
       \textbf{DI} (previous)  &  &  & \\
       \textbf{DBI} (ours)  & \checkmark &  & \\
       \textbf{DDBI} (ours) & \checkmark & \checkmark & \\
       \textbf{DSBI} (ours) & \checkmark & \checkmark & \checkmark\\
       \hline
    \end{tabular}}
    % \vskip -0.1in
    \label{tab:my_label}
\end{table}

The main contributions of this work are as follows: 
\begin{compactitem}%\setlength{\itemsep}{1pt}
    \item We propose $\text{D}^3\text{RE}$, the first unified framework to address both the density-chasm and support-chasm problems via uniformly approximated density ratio estimation (\cref{proposition:uniform-estimation-formula}). Our interpolants expand both the support (\cref{theorem:support-set-expansion}) and trajectory sets (\cref{corollary:path-set-expansion}),  alleviating the support-chasm problem.
    
    \item We incorporate guidance mechanisms to improve interpolant quality: GD improves the stability of time score functions at boundary (\cref{theorem:divergent-lower-bound-time-score},\cref{corollary:upper-bound-time-score-DDBI-DSBI}), while OTR leads to more stable (\cref{theorem:upper-bound-of-DDBI-DSBI}) and efficient (\cref{fig:nfe-comparison-density-ratio-toy2toy}) estimation.
    
    \item Experiments demonstrate  $\text{D}^3\text{RE}$'s superiority in density ratio estimation, mutual information estimation, and density estimation. 
\end{compactitem}

% \vspace{-3mm}

\section{Background}

Let $\mathbf{X}_0\sim q_0(\mathbf{x})$ and $\mathbf{X}_1\sim q_1(\mathbf{x})$ be random variables in $\mathbb{R}^d$ with set-theoretic supports $\mathsf{supp}(q_0)$ and $\mathsf{supp}(q_1)$. Upper cases $\mathbf{X}_0$ and $\mathbf{X}_1$ denote random variable, while lower cases $\mathbf{x}_0$ and $\mathbf{x}_1$ denote the samples of random variables. 

% \vspace{-3mm}

\subsection{Density Ratio Estimation}
Density ratio estimation (DRE) aims to estimate the density ratio $r^{\star}(\mathbf{x}) = \frac{q_1(\mathbf{x})}{q_0(\mathbf{x})}$ using i.i.d.\ samples from both distributions. 
A common approach is density ratio matching via Bregman divergence minimization  \citep{sugiyama2012density}, which optimizes a parameterized ratio $r_{\boldsymbol{\theta}}$ by minimizing
% \resizebox{0.95\linewidth}{!}{
% \begin{equation}
%     \mathrm{BD}(r_{\boldsymbol{\theta}})=-\mathbb{E}_{q_0(\mathbf{x}_0)\atop q_1(\mathbf{x}_1)}\left[ \log\frac{1}{1+r_{\boldsymbol{\theta}}(\mathbf{x}_0)} +\log\frac{r_{\boldsymbol{\theta}}(\mathbf{x}_1)}{1+r_{\boldsymbol{\theta}}(\mathbf{x}_1)} \right], 
%        \label{eq:logistic-regression}
% \end{equation}
% }
\begin{equation}
\resizebox{0.9\linewidth}{!}{%
        $\mathrm{BD}(r_{\boldsymbol{\theta}})=-\mathbb{E}_{\substack{q_0(\mathbf{x}_0)\\q_1(\mathbf{x}_1)} }\left[ \log\frac{1}{1+r_{\boldsymbol{\theta}}(\mathbf{x}_0)} +\log\frac{r_{\boldsymbol{\theta}}(\mathbf{x}_1)}{1+r_{\boldsymbol{\theta}}(\mathbf{x}_1)} \right]$, 
}
\label{eq:logistic-regression}
\end{equation}
where $\boldsymbol{\theta}$ denotes the trainable parameters of  $r_{\boldsymbol{\theta}}$.
The minimizer of Eq.\ (\ref{eq:logistic-regression}) satisfies $r_{\boldsymbol{\theta}^{\star}}=r^{\star}$. However, DRE suffers from the density-chasm problem \citep{rhodes2020telescoping}.

TRE mitigates this by employing a \textit{divide-and-conquer} strategy. It partitions the interval $[0,1]$ into $M\in\mathbb{Z}_{+}$ sub-intervals with endpoints $\{m/M\}_{m=0}^{M}$, constructing intermediate variables $\mathbf{X}_{m/M}\sim q_{m/M}(\mathbf{x})$ via linear interpolation
\begin{equation}
    \mathbf{X}_{m/M} = \sqrt{1-\eta_{m/M}^{2}}\mathbf{X}_0 + \eta_{m/M} \mathbf{X}_1,   \label{eq:interpolant-TRE}
\end{equation}
where $\eta_{m/M}$ increases from 0 to 1. The density ratio decomposes into a telescoping product
\begin{equation}
\begin{aligned}
    r^{\star}(\mathbf{x})=\frac{q_{1}(\mathbf{x})}{q_0(\mathbf{x})}&=\prod_{m=0}^{M-1} \frac{q_{(m+1)/M}(\mathbf{x})}{q_{m/M}(\mathbf{x})}\\
    &=\prod_{m=0}^{M-1}r_{m/M}^{\star}(\mathbf{x})\approx \prod_{m=0}^{M-1}r_{\boldsymbol{\theta}_m}(\mathbf{x}),
\end{aligned} \label{eq:density-ratio-TRE}
\end{equation}
where $r_{m/M}^{\star}$ is the target intermediate density ratio, estimated by a parameterized neural network $r_{\boldsymbol{\theta}_m}$ with trainable parameters $\boldsymbol\theta_m$. 
In this case, $M$  networks must be trained. While a larger $M$ improves accuracy, it increases computational cost and may still fail to sufficiently reduce the KL divergence, $ \mathrm{KL}(q_{m/M}\|q_{(m+1)/M}) $, leaving the density-chasm problem unaddressed. 

DRE-$ \infty $ \citep{choi2022density} extends TRE to $ M \to \infty $, representing the log ratio as an integral of the time score
\begin{equation}
 \log r^{\star}(\mathbf{x})=\int_{0}^{1}\partial_t\log q_{t}(\mathbf{x}) \mathrm{d}t\approx \int_0^1 s_{\boldsymbol{\theta}^\star}^t (\mathbf{x}, t)\mathrm{d}t,
\end{equation}
where $\partial_t\log q_{t}(\mathbf{x})\approx (\log q_{t+\Delta t}(\mathbf{x}) - \log q_{t}(\mathbf{x}))/\Delta t$ with an infinitesimal gap $\Delta t=\lim_{M\to\infty} 1/M$ denotes the time score. 
The time score model $s_{\boldsymbol{\theta}}^t$ approximates the time score via minimization of
\begin{equation}
\mathcal{L}_{1}=\mathbb{E}_{q(t)q_{t}(\mathbf{x})}\left[\lambda(t)\left|\partial_t\log q_{t}(\mathbf{x})-s_{\boldsymbol{\theta}}^{t}(\mathbf{x},t)\right|^{2}\right],    \label{eq:time-score-matching}   % 
\end{equation}
where $\lambda(\cdot) : [0,1]\to\mathbb{R}_{+}$ is a time-dependent weighting function, and $q(t)=\mathcal{U}[0,1]$ is the uniform distribution over the interval [0,1]. 
When the time score model satisfies $\partial_t\log q_{t}(\mathbf{x})=s_{\boldsymbol{\theta}^{\star}}^{t}(\mathbf{x},t)$, the loss function reaches its minimum value $\mathcal{L}_{1}(\boldsymbol{\theta}^{\star})$. Unlike in TRE, only one network, $s_{\boldsymbol{\theta}}^{t}$, needs to be trained in DRE-$\infty$. 

\subsection{Denoising Diffusion Bridge Model}
Denoising diffusion models (DDMs) simulate a diffusion process $\{\mathbf{X}_t\}_{t\in[0,1]}$, which serves as a continuous bridge between $\mathbf{X}_0$ and $\mathbf{X}_1$. 
This process is described by the solution to an Itô stochastic differential equation (SDE)
\begin{equation}
	\mathrm{d} \mathbf{X}_t=\mathbf{f}(\mathbf{X}_t,t)\mathrm{d}t+g(t)\mathrm{d}\mathbf{W}_t, \label{eq:sde} 
\end{equation}
where $\{\mathbf{W}_t\}_{t\in[0,1]}$ is a standard Wiener process, $\mathbf{f}: \mathbb{R}^d\times [0,1] \to \mathbb{R}^d$ and $g: [0,1] \to \mathbb{R}$ are termed as the drift coefficient and scalar diffusion coefficient of $\mathbf{X}_t$, respectively. 

Conventional DDMs \citep{li2023scire,song2020score,li2024neural,xu2024sparse} require either $q_0$ or $q_1$ to be a simple, tractable distribution (e.g., isotropic Gaussian), which limits their ability to bridge arbitrary complex distributions and restricts applications such as DRE.

Denoising diffusion bridge models  \citep{zhou2023denoising} overcome this limitation by simulating stochastic processes that interpolate between paired distributions with $\mathbf{X}_1$ as endpoints. 
These processes are derived from the SDE in Eq. (\ref{eq:sde}) via Doob's $h$-transform \citep{doob1984classical},
\begin{equation}
	\mathrm{d} \mathbf{X}_t=[\mathbf{f}(\mathbf{X}_t,t)+g^2(t)\mathbf{h}(\mathbf{X}_t,t,\mathbf{X}_1)]\mathrm{d}t+g(t)\mathrm{d}\mathbf{W}_t, \label{eq:sde-bridge} 
\end{equation}
where 
% $\tilde{\mathbf{f}}(\mathbf{X}_t,t,\mathbf{X}_1)=\mathbf{f}(\mathbf{X}_t,t)+g^2(t)\mathbf{h}(\mathbf{X}_t,t,\mathbf{X}_1)$ denotes the shifted drift and 
$\mathbf{h}(\mathbf{X}_t,t,\mathbf{X}_1)=\nabla_{\mathbf{x}_t}\log p_t(\mathbf{X}_1\mid \mathbf{X}_t)$ is the $h$ function representing the gradient of the log transition kernel from time $t$ to 1. 
The process explicitly depends on $\mathbf{X}_1$ and, given $\mathbf{X}_0$ and $\mathbf{X}_1$, it forms a \textit{diffusion bridge} with transition kernel $q_{t}(\mathbf{x}\mid \mathbf{X}_0,\mathbf{X}_1)$. 
A special case of the diffusion bridge, termed the \textit{Brownian bridge}, arises under the conditions $\mathbf{f}(\mathbf{X}_t,t):=0, g(t):=1$ and $\mathbf{h}(\mathbf{X}_t,t,\mathbf{X}_1):=\frac{\mathbf{X}_1-\mathbf{X}_t}{1-t}$. 
The Brownian bridge is defined as the solution to $\mathrm{d}\mathbf{B}_t=\frac{\mathbf{B}_1-\mathbf{B}_t}{1-t}\mathrm{d}t+\mathrm{d}\mathbf{W}_t$ and its transition kernel is given by
$q_t(\mathbf{b}\mid\mathbf{B}_0,\mathbf{B}_1)=\mathcal{N}((1-t) \mathbf{B}_0 + t\mathbf{B}_1, t(1-t)\mathbf{E}_d)$.

\section{Method}
In this section, we we extend prior works into a unified framework called \textbf{D}equantified \textbf{D}iffusion-bridge \textbf{D}ensity-\textbf{R}atio \textbf{E}stimation (\textbf{$\text{D}^3\text{RE}$}). $\text{D}^3\text{RE}$ offers a robust and efficient solution for density ratio estimation, theoretically mitigating the density-chasm and support-chasm problems. 

\subsection{Support-chasm Problem}

\paragraph{Deterministic Interpolant.} 
We summarize the interpolants used in prior works, such as TRE and DRE-$\infty$, to \textit{deterministic interpolant} (\textbf{DI}), defined as
\begin{equation}%[box=\fbox]{equation}
    \mathbf{I}(\mathbf{X}_0,\mathbf{X}_1,t)=\alpha_t\mathbf{X}_0+\beta_t\mathbf{X}_1, \label{eq:deterministic-interpolant}  %\mathbf{X}_t=
\end{equation}
where $\mathbf{I}: \mathbb{R}^d\times\mathbb{R}^d\times [0,1]\to\mathbb{R}^d$ is an interpolant  continuously differentiable in $(\mathbf{X}_0, \mathbf{X}_1, t)$ and the time-indexed coefficients $\alpha_t,\beta_t\in C^2 [0,1]$ are monotonic with $\alpha_t$ decreasing and $\beta_t$ increasing in $t$. They are strictly positive and  satisfy boundary conditions $\alpha_0=\beta_1=1$ and $\alpha_1=\beta_0=0$, with constraints $\alpha_t+\beta_t=1$ or $\alpha_t^2+\beta_t^2=1, \forall t\in[0,1]$. 

% The process $\{\mathbf{X}_t\}_{t\in[0,1]}$ associated with Eq.\ (\ref{eq:deterministic-interpolant}), is called the \textit{deterministic interpolant} (DI). 
Prior methods, such as TRE \citep{rhodes2020telescoping} and DRE-$\infty$ \citep{choi2022density}, are specific cases of DI distinguished by their choices of $\alpha_t$ and $\beta_t$ (see \cref{appendix:special-cases-DI}). 

However, DRE with DI suffers from the support-chasm problem, where minimal overlap between $\mathsf{supp}(q_0)$ and $\mathsf{supp}(q_1)$ leads to ill-defined ratios \citep{srivastava2023estimating}. 

\begin{definition}[Support-chasm Problem]\label{definition:support-chasm-problem}
   Let $ q_0, q_1 $ be probability density functions with supports $ \mathsf{supp}(q_0) $ and $ \mathsf{supp}(q_1) $, respectively. For a given threshold $ \varepsilon > 0 $, if $\mu\left( \mathsf{supp}(q_0) \cap \mathsf{supp}(q_1) \right) < \varepsilon$, then a \textit{support-chasm} exists between $ q_0 $ and $ q_1 $, where $ \mu $ is the Lebesgue measure.
\end{definition}

\paragraph{Diffusion Bridge Interpolant.} 
To mitigate the support-chasm, we introduce a Brownian bridge $\{\mathbf{B}_t\}_{t\in[0,1]}$, leading to the \textit{diffusion bridge interpolant} (\textbf{DBI})
\begin{equation} \label{eq:diffusion-bridge-interpolant-bridge}
    \mathbf{X}_t=\mathbf{I}(\mathbf{X}_0,\mathbf{X}_1,t) +\gamma\mathbf{B}_t,
\end{equation}
where $\gamma\in\mathbb{R}_{\geq0}$ is the noise factor controlling the stochastic component $\mathbf{B}_t$.
This factor provides flexibility by adjusting the variability introduced by the bridge at different stages of interpolation.
\textit{When $\gamma=0$, DBI reduces to DI. }

Since $\mathbf{B}_t$ is a Gaussian process with zero mean and variance $t(1-t)$, i.e., $\mathbf{B}_t\sim\mathcal{N}(\mathbf{0},t(1-t)\mathbf{E}_d)$, the transition kernel of the DBI can be derived follows from \cref{eq:diffusion-bridge-interpolant-bridge} as  $q_t(\mathbf{x}\mid\mathbf{X}_0,\mathbf{X}_1)=\mathcal{N}(\mathbf{I}(\mathbf{X}_0,\mathbf{X}_1,t),t(1-t)\gamma^2\mathbf{E}_d)$. By applying the reparameterization trick, DBI admits the equivalent form
\begin{equation}
    \mathbf{X}_t=\mathbf{I}(\mathbf{X}_0,\mathbf{X}_1,t) +\sqrt{ t(1-t)\gamma^2}\mathbf{Z}_t,    \label{eq:diffusion-bridge-interpolant}
\end{equation}
where $\mathbf{Z}_t\sim\mathcal{N}(\mathbf{0},\mathbf{E}_d)$, ensuring analytical tractability and efficient implementation. 
The term $\sqrt{t(1-t)\gamma^2}\mathbf{Z}_t$ adds controlled variability and provides robustness and flexibility to the interpolant, expanding the support of $q_t$.

\begin{theorem}[Support Set Expansion]
\label{theorem:support-set-expansion}
    Let $\mathbf{X}_0$ and $\mathbf{X}_1$ be random variables.
    Let $q_t(\mathbf{x})$ and $q_t^\prime(\mathbf{x})$ denote the marginal densities under DI and DBI, respectively.
    Then, for any $t\in(0,1)$, the support of $q_{t}^\prime$ includes or expands beyond the support of $q_t$, i.e.,  $\mathsf{supp}(q_t^\prime) \supseteq \mathsf{supp}(q_t)$. 
\end{theorem}
See \cref{proof:support-set-expansion} for detailed derivation.
This result shows that DBI covers a larger or equal region of the space compared to DI, providing theoretical justification for its ability to mitigate the support-chasm problem in $\text{D}^3\text{RE}$.

\begin{corollary}[Trajectory Set Expansion]  \label{corollary:path-set-expansion}
    Under the same setup as in \cref{theorem:support-set-expansion}, let the trajectory sets generated by the DI and the DBI be denoted by $\mathbb{T}=\{\{\mathbf{x}_t\}_{t\in[0,1]}; \mathbf{x}_t\in\mathsf{supp}(q_t)\}$ and $\mathbb{T}^{\prime}=\{\{\mathbf{x}_t^{\prime}\}_{t\in[0,1]}; \mathbf{x}_t^{\prime}\in\mathsf{supp}(q_t^{\prime})\}$, respectively. Then, $\mathbb{T}^{\prime}$ contains $\mathbb{T}$, i.e., $\mathbb{T}^{\prime}\supseteq\mathbb{T}$.
\end{corollary}
See \cref{proof:path-set-expansion} for details. 
This generalizes the support expansion result to entire paths, implying that DBI, by injecting noise, explores a broader set of trajectories than DI.
As a result, it provides better coverage of the interpolation space, enhancing robustness across diverse distributions.

% \begin{corollary}\label{corollary:reduced-estimation-error}
%     Under the setup in \cref{theorem:support-set-expansion}, DDBI reduces bias in estimating  $r^{\star}(\mathbf{x})$ for  $\mathbf{x}\in\mathsf{supp}(q_0)\cap\mathsf{supp}(q_1)$, with a negligible bias increase when $\gamma^2$ is small. %\textcolor{red}{controlled and negligible bias, $\mathcal{O}(\gamma^2)$}
% \end{corollary}

% See \cref{proof:reduced-estimation-error} for the detailed derivation. 
% This highlights that DDBI improves estimation  accuracy by bridging support gaps and expanding the trajectory space.

\subsection{Gaussian Dequantization for Mollified Time Score}

A second fundamental challenge in conventional interpolants such as DI and DBI is the divergence of the absolute time score, as shown in \cref{theorem:divergent-lower-bound-time-score}.

\begin{theorem}  \label{theorem:divergent-lower-bound-time-score}
    Let $\{\mathbf{X}_t\}_{t\in[0,1]}$ be a DI defined in \cref{eq:deterministic-interpolant}. 
    Under \cref{ass:smoothness} and \cref{assumption:Lip-conditional-distri}, the time score for any $t\in(0,1)$ satisfies
    \begin{equation}
    \begin{aligned}
         \mathbb{E}_{q_t}[|\partial_t \log q_t|] &\geq d\left( (1-L)\frac{|\dot{\alpha}_t|}{\alpha_t} - L\frac{|\dot{\beta}_t|}{\beta_t} \right) \\
         &- \mathcal{O}\left(\sqrt{ \mathbb{E}_{q_t}[\|\nabla \log q_t\|^2] }\right),
    \end{aligned}
    \end{equation}
where $L$ is the Lipschitz constant in \cref{assumption:Lip-conditional-distri}.  
Moreover, if $L<1$, this lower bound diverges to  infinity
\begin{equation}
\begin{aligned}
    % \lim_{t \to 0^+} \mathbb{E}_{q_t}\left[|\partial_t \log q_t|\right] &= -\infty, \\ 
    \lim_{t \to 1^-} \mathbb{E}_{q_t}\left[|\partial_t \log q_t|\right] &= +\infty.  
\end{aligned}
\end{equation}
\end{theorem}

Proofs can be found in \cref{proof:divergent-lower-bound-time-score}.

\paragraph{Dequantified Diffusion Bridge Interpolant.}
To stabilize the time score near the boundary, we introduce Gaussian dequantization (GD) by adding controlled perturbations to boundary samples, effectively handling the boundary densities and resulting uniformly bounded time score across $[0,1]$ (see \cref{corollary:upper-bound-time-score-DDBI-DSBI} for details).

Specifically, for $\mathbf{x}_i \sim q_i$, its dequantified form $\mathbf{x}_i^\prime$ can be obtained by
\begin{equation}
\mathbf{x}_i^\prime = \mathbf{x}_i + \mathbf{z}_{\varepsilon}, \quad \mathbf{z}_{\varepsilon} \sim \mathcal{N}(\mathbf{0}, \varepsilon \mathbf{E}_d), \quad i \in \{0,1\},
\end{equation}
where $\varepsilon\in\mathbb{R}_{+}$ is small. The resulting dequantified densities are obtained via Gaussian convolution, $q_i^\prime = q_i \ast  \mathcal{N}(\mathbf{0}, \varepsilon \mathbf{E}_d)$. 
This smoothing ensures bounded time scores near $t=0$ and $t=1$ (see \cref{theorem:upper-bound-of-DDBI-DSBI} and  \cref{corollary:upper-bound-time-score-DDBI-DSBI} for details), improving  stability in DRE. 

Incorporating GD into the DBI yields the \textit{dequantified diffusion bridge interpolant} (\textbf{DDBI}), formulated as
\begin{equation} % [box=\fbox]{equation}
    \mathbf{X}_t^\prime = \mathbf{I}(\mathbf{X}_0^\prime, \mathbf{X}_1^\prime, t) + \sqrt{t(1-t)\gamma^2} \mathbf{Z}_t.
\end{equation}
The DDBI can be expressed in terms of the original DBI by defining perturbed variables as $\mathbf{X}_i^\prime = \mathbf{X}_i + \mathbf{Z}_\varepsilon$, where $\mathbf{Z}_\varepsilon \sim \mathcal{N}(\mathbf{0}, \varepsilon \mathbf{E}_d)$.
This results in
\begin{equation}
\mathbf{X}_t^\prime = \mathbf{I}(\mathbf{X}_0, \mathbf{X}_1, t) + \sqrt{t(1-t)\gamma^2 + (\alpha_t^2 + \beta_t^2) \varepsilon} \mathbf{Z}_t.
\end{equation}
Here, the additional term $(\alpha_t^2 + \beta_t^2)\varepsilon$ reflects the effect of GD, yielding smoother interpolation. As a result, the transition kernel of the DDBI, $q_t^\prime(\mathbf{x} \mid \mathbf{x}_0, \mathbf{x}_1)$, is given by $\mathcal{N}(\mathbf{I}(\mathbf{x}_0,\mathbf{x}_1,t), \left(t(1-t)\gamma^2 + (\alpha_t^2  + \beta_t^2) \varepsilon\right)\mathbf{E}_d)$, which shares the same mean trajectory as DBI (see \cref{eq:diffusion-bridge-interpolant}).
% At the boundaries, the transition kernels  shift from Dirac delta distributions $\delta(\mathbf{X}_i-\mathbf{x}_i)$ to Gaussian distributions $ \mathcal{N}(\mathbf{x}_i, \varepsilon \mathbf{E}_d)$. 
% This ensures regularity and stability while enabling flexible interpolation in sparse data spaces \citep{liu2019neural}.

We have also analyzed the uniform approximation of density ratio using the DDBI.
The relationship between $ r(\mathbf{x})$ and $ r^\prime(\mathbf{x}) = \frac{q_1^\prime(\mathbf{x})}{q_0^\prime(\mathbf{x})}$ is characterized by  \cref{proposition:uniform-estimation-formula}.
\begin{proposition}
\label{proposition:uniform-estimation-formula}
    Let $r(\mathbf{x})$ and $r^\prime(\mathbf{x})$ be the density ratios with and without GD, respectively. Then, $r^\prime$ is a uniform apporximation of $r$, with the error bounded by:
    \begin{equation}
        \| r^\prime - r \|_{L^\infty} \leq \mathcal{O}(\varepsilon).
    \end{equation}
    Thus, as $ \varepsilon \to 0 $, $ r^\prime(\mathbf{x}) \to r(\mathbf{x}) $ in the uniform norm. 
\end{proposition}
See \cref{proof:uniform-estimation-formula} for detailed derivation.
\cref{proposition:uniform-estimation-formula} confirms that the dequantified density ratio $r^\prime(\mathbf{x})$ is a \textit{uniform approximation} of $r(\mathbf{x})$ for sufficiently small $\varepsilon$. 

\subsection{Optimal Transport Rearrangement} 
To further reduce estimation error and improve efficiency of DRE, we extend the probability path of DDBI by aligning it with the entropically regularized optimal transport (OT)
\begin{equation} 
    \pi_{2\gamma^2} = \underset{\pi \in \Pi(q_0^\prime,q_1^\prime)}{\text{argmin}} \int \|\mathbf{x}_0 - \mathbf{x}_1\|^2 \mathrm{d}\pi - 2\gamma^2 \mathcal{H}(\pi),  \label{eq:entropic-OT-problem}
\end{equation} 
where $\Pi(q_0^\prime, q_1^\prime)$ is the set of all probability paths with marginals $q_0^\prime$ and $q_1^\prime$, and $\mathcal{H}(\pi)$ is the entropy of $\pi$. The coefficient $2\gamma^2$ is regularization factor (details in \cref{appendix:optimal-transport}). 
% \textcolor{red}{include Hamilton MC}

We apply the scalable Sinkhorn algorithm \citep{cuturi2013sinkhorn} to mini-batches $\{\left(\mathbf{x}_{0}^\prime, \mathbf{x}_{1}^\prime\right)_{n}\}_{n=1}^{N}\sim q_0^\prime\times q_1^\prime$, obtaining   rearranged sample pairs $\{\left(\hat{\mathbf{x}}_{0}^\prime, \hat{\mathbf{x}}_{1}^\prime\right)_n \}_{n=1}^{N}{\sim}\pi_{2\gamma^2}$. For convenience, we refer to this procedure as \emph{optimal transport rearrangement} (OTR), which preserves the marginals and only changes the sample pairing.

\paragraph{Dequantified Schr{\"o}dinger Bridge Interpolant.}
Applying OTR followed by DDBI yields the \textit{dequantified Schr{\"o}dinger bridge interpolant} (\textbf{DSBI})
\begin{equation}%[box=\fbox]{equation}
    \hat{\mathbf{X}}_t^\prime = \mathbf{I}(\hat{\mathbf{X}}_0^\prime, \hat{\mathbf{X}}_1^\prime, t) + \sqrt{t(1-t)\gamma^2 + (\alpha_t^2 + \beta_t^2) \varepsilon} \mathbf{Z}_t,  \label{eq:DSBI}
\end{equation}
where $\alpha_t = 1 - t$ and $\beta_t = t$ are fixed.

% $\mathbf{I}(\hat{\mathbf{X}}_0^\prime, \hat{\mathbf{X}}_1^\prime, t)=(1-t)\hat{\mathbf{X}}_0^\prime + t\hat{\mathbf{X}}_1^\prime$.

We show that rearranging the mini-batches via OTR leads to an interpolant that naturally solves the  Schr{\"o}dinger bridge (SB) problem \citep{schrodinger1932theorie}.
\begin{proposition}
    \label{proposition:solution-to-SB-problem}
    The probability path defined by DSBI solves the SB problem
    \begin{equation}
    \pi^{\star} = \underset{\pi \in \Pi(q_0^\prime,q_1^\prime)}{\text{argmin}}  \mathrm{KL}(\pi \| \pi_{\mathrm{ref}}),    \label{eq:SB-problem}
    \end{equation}
    where $\pi_{\mathrm{ref}}$ is a Wiener process scaled by $\gamma$.
\end{proposition}
See \cref{proof:solution-to-SB-problem} for  proof. This result suggests that DSBI, as a principled integration of DDBI and OTR, implicitly solves the SB problem and provides a minimum-cost stochastic interpolation between $q_0^\prime$ and $q_1^\prime$.

Furthermore, to rigorously quantify the improvement brought by OTR, we establish the following result comparing the upper error bounds of DDBI and DSBI.

\begin{theorem}\label{theorem:upper-bound-of-DDBI-DSBI}
    Consider the DDBI and DSBI  with  $\alpha_t = 1-t$, $\beta_t = t$. Let $\pi \in \Pi(q_0^\prime, q_1^\prime)$ be any coupling for DDBI, and  $\pi_{2\gamma^2}$ the entropically regularized  OT coupling for DSBI. 
    Then, for all $t\in[0,1]$, the variance of the time score under DSBI is no greater than that under DSBI, i.e.,
    \begin{equation}
        \mathrm{Var}^{\text{DSBI}}_{q_t^\prime}(\partial_t \log q_t^\prime)\le \mathrm{Var}^{\text{DDBI}}_{q_t^\prime}(\partial_t \log q_t^\prime).
    \end{equation}
    % Define the error functional $\mathcal{E} := \left| \int_0^1 \partial_t \log q_t^\prime \mathrm{d}t - \log \frac{q_1^\prime}{q_0^\prime} \right|$.
    % Then, $\mathbb{E}_{\pi_{2\gamma^2}}[\mathcal{E}_{\text{DSBI}}^2]$ has a tighter error upper bound than $\mathbb{E}_{\pi}[\mathcal{E}_{\text{DDBI}}^2]$.
\end{theorem}

\begin{corollary} \label{corollary:upper-bound-time-score-DDBI-DSBI}
For all $t \in [0,1]$, the time score of DDBI is uniformly bounded by
\begin{equation}
    \mathbb{E}_{q_t^\prime}[|\partial_t \log q_t^\prime|] \leq \sqrt{\frac{1}{\sigma_t^2}\mathbb{E}_{\pi}\left[\left\|\mathbf{X}_0^\prime - \mathbf{X}_1^\prime\right\|^2 \right] + \frac{\dot{\sigma}_t^4 d}{2\sigma_t^4}},
\end{equation}
where $\sigma_t^2 = t(1-t)\gamma^2 + (2t^2-2t+1)\varepsilon$ is strictly positive.
\end{corollary}
See \cref{proof:upper-bound-of-DDBI-DSBI} and \cref{proof:upper-bound-time-score-DDBI-DSBI} for detailed derivations of \cref{theorem:upper-bound-of-DDBI-DSBI} and \cref{corollary:upper-bound-time-score-DDBI-DSBI}. These results establish that both DDBI and DSBI admit a bounded time-integrated time score,  $\mathbb{E}_{q_t^\prime}[|\partial_t \log q_t^\prime|]$, in contrast to the divergent lower bound exhibited by DI (see \cref{theorem:divergent-lower-bound-time-score}). Moreover, it provides a formal justification for the error reduction achieved by DSBI through the coupling $\pi_{2\gamma^2}$.

\subsection{Dequantified Diffusion Bridge DRE} 

For the DDBI, $r^\prime(\mathbf{x})$ can be  approximated effectively  using a neural network, as formulated in \cref{proposition:log-density-ratio}. See \cref{proof:theorem-log-density-ratio} for detailed derivation. 
\begin{definition}[Dequantified Diffusion bridge Density Ratio Estimation, D$^3$RE]  
\label{proposition:log-density-ratio} 
Given the marginal probability density of DDBI, $q_{t}^\prime(\mathbf{x})$, the log density ratio for a given point  $\mathbf{x}\in\mathbb{R}^{d}$can be estimated as
\begin{equation}
    \log r^{\star} (\mathbf{x}) \approx \int^{1}_{0} \partial_t \log q_{t}^\prime(\mathbf{x}) \mathrm{d}t, \label{eq:log-density-ratio-true}
\end{equation}
where $\partial_t\cdot$ denotes the time derivative operator.
\end{definition}

\textbf{Time Score-matching Loss.} We train a time score model $s_{\boldsymbol{\theta}}^{t}(\mathbf{x},t)$ to approximate the time score $\partial_t \log q_{t}^\prime(\mathbf{x})$ by minimizing the  time score-matching loss \citep{choi2022density},
\begin{equation}
	\mathcal{L}_{2}=\mathbb{E}_{ q(t) q_t^\prime(\mathbf{x})}\bigg[\lambda(t)\left|\partial_t\log q_t^\prime(\mathbf{x}) - s_{\boldsymbol{\theta}}^{t}(\mathbf{x},t)\right|^{2}\bigg].    \label{eq:time-score-matching-bridge}
\end{equation}
However, $\partial_t\log q_t^\prime(\mathbf{x}) $ is intractable in practice. 
To bypass this, an equivalent integration-by-parts form \citep{song2020improved, choi2022density} is proposed
\begin{equation}
	\begin{aligned}
		&\mathcal{L}_{3}=\mathbb{E}_{ q_{0}^\prime(\mathbf{x}_0) q_{1}^\prime(\mathbf{x}_1)}[\lambda(0)s_{\boldsymbol{\theta}}^{t}(\mathbf{x}_0,0)-\lambda(1)s_{\boldsymbol{\theta}}^{t}(\mathbf{x}_1,1)] \\
        &+\mathbb{E}_{ q(t) q_t^\prime(\mathbf{x})}\left[\partial_t\left[\lambda(t)s_{\boldsymbol{\theta}}^{t}(\mathbf{x},t)\right]+\frac{1}{2}\lambda(t)s_{\boldsymbol{\theta}}^{t}(\mathbf{x},t)^2\right],
	\end{aligned}   \label{eq:time-score-matching-loss-L4}
\end{equation}
where $\partial_t\left[\lambda(t)s_{\boldsymbol{\theta}}^{t}(\mathbf{x},t)\right]=\lambda(t)\partial_ts_{\boldsymbol{\theta}}^{t}(\mathbf{x},t)+\lambda^{\prime}(t)s_{\boldsymbol{\theta}}^{t}(\mathbf{x},t)$, $\partial_ts_{\boldsymbol{\theta}}^{t}(\mathbf{x},t)$ and $\lambda^{\prime}$ denote the time derivative of the time score model and weighting function, respectively. The first two terms  enforce the boundary conditions. 
$\mathcal{L}_2$ and $\mathcal{L}_3$ differ only by a constant $C$ independent of $\boldsymbol{\theta}$. 
In practice, for stable and effective training, the joint score-matching loss is implemented, as described in \cref{appendix:joint-score-matching}.

\paragraph{Estimating Target Log Density Ratio.} 
Given the optimal parameters $\boldsymbol{\theta}^\star$ obtained by minimizing $\mathcal{L}3$, the log density ratio at any point $\mathbf{x}$ can be estimated as
\begin{equation}
    \log r^{\star} (\mathbf{x}) \approx \int^{1}_{0} \partial_t \log q_{t}^\prime(\mathbf{x}) \mathrm{d}t\approx \int^{1}_{0}s_{\boldsymbol{\theta}^{\star}}^{t}(\mathbf{x},t) \mathrm{d}t,  \label{eq:logr-estimation}
\end{equation}
based on \cref{proposition:log-density-ratio}.
See \cref{algorithm:DDBI,algorithm:DSBI} for the full training and estimation procedures using DDBI and DSBI.

% \vspace{-3mm}

\begin{algorithm}
	% \setstretch{1.20}
    \renewcommand{\algorithmicrequire}{\textbf{Input:}}
    \renewcommand{\algorithmicensure}{\textbf{Output:}}
    \caption{Training and estimation of $\text{D}^3\text{RE}$ with DDBI}
    \label{algorithm:DDBI}
\begin{algorithmic}
    \REQUIRE Probability densities $q_0$ and $q_1$, time score model $s_{\boldsymbol{\theta}}^{t}$, coefficients $\alpha_t$ and $\beta_t$, noise factor $\gamma$ and $\varepsilon$.
    \STATE \textbf{Initialize:} trainable parameters $\boldsymbol{\theta}$ of $s_{\boldsymbol{\theta}}^{t}$, a given point $\mathbf{x}$. 
    \STATE $\mathbf{x}_0\sim q_0(\mathbf{x}), \mathbf{x}_1\sim q_1(\mathbf{x}), t\sim \mathcal{U}(0,1)$
    \STATE $ \mathbf{z}_{\varepsilon}\sim\mathcal{N}(\mathbf{0},\varepsilon\mathbf{E}_d),\mathbf{z}\sim\mathcal{N}(\mathbf{0},\mathbf{E}_d)$
    \STATE $\mathbf{x}^{\prime}_0 \leftarrow \mathbf{x}_0 + \mathbf{z}_{\varepsilon}, \mathbf{x}^{\prime}_1 \leftarrow \mathbf{x}_1 + \mathbf{z}_{\varepsilon}$      \hfill  $\%\ \text{GD}~~~~$
    \STATE $\mathbf{x}_t^\prime\leftarrow \alpha_t \mathbf{x}_0^\prime+\beta_t\mathbf{x}_1^\prime +\sqrt{t(1-t)\gamma^2}\mathbf{z}$ \hfill $\% \ \text{DBI}\,~~$
    \STATE $\boldsymbol{\theta}^{\star}\leftarrow\mathrm{Adam}(\boldsymbol{\theta},\nabla_{\boldsymbol{\theta}}\mathcal{L}_3(\boldsymbol{\theta}))$
    \STATE $\log r(\mathbf{x})\leftarrow \mathsf{odeint\_adjoint}(s_{\boldsymbol{\theta}^{\star}}^{t},(0,1),\mathbf{x})$
    \ENSURE estimated log density ratio $\log r(\mathbf{x})$.
\end{algorithmic}
\end{algorithm}

% \vspace{-5mm}
 
\begin{algorithm}
	% \setstretch{1.20}
    \renewcommand{\algorithmicrequire}{\textbf{Input:}}
    \renewcommand{\algorithmicensure}{\textbf{Output:}}
    \caption{Training and estimation of $\text{D}^3\text{RE}$ with DSBI}
    \label{algorithm:DSBI}
\begin{algorithmic}
    \REQUIRE Probability densities $q_0$ and $q_1$, time score model $s_{\boldsymbol{\theta}}^{t}$, coefficients $\alpha_t$ and $\beta_t$, noise factor $\gamma$ and $\varepsilon$.
    \STATE \textbf{Initialize:} trainable parameters $\boldsymbol{\theta}$ of $s_{\boldsymbol{\theta}}^{t}$, a given point $\mathbf{x}$. 
    \STATE $\mathbf{x}_0\sim q_0(\mathbf{x}), \mathbf{x}_1\sim q_1(\mathbf{x}), t\sim \mathcal{U}(0,1)$
    \STATE $ \mathbf{z}_{\varepsilon}\sim\mathcal{N}(\mathbf{0},\varepsilon\mathbf{E}_d),\mathbf{z}\sim\mathcal{N}(\mathbf{0},\mathbf{E}_d)$
    \STATE $\mathbf{x}^{\prime}_0 \leftarrow \mathbf{x}_0 + \mathbf{z}_{\varepsilon}, \mathbf{x}^{\prime}_1 \leftarrow \mathbf{x}_1 + \mathbf{z}_{\varepsilon}$      \hfill  $\%\ \text{GD}~~~~$
    
    \STATE $\pi_{2\gamma^2}\leftarrow \mathsf{Sinkhorn}(\mathbf{x}_0^{\prime}, \mathbf{x}_1^{\prime},2\gamma^2)$  \hfill $\%\ \text{OTR}~~$
    \STATE $(\hat{\mathbf{x}}_0^{\prime}, \hat{\mathbf{x}}_1^{\prime})\sim \pi_{2\gamma^2}$
    \STATE $\hat{\mathbf{x}}^\prime\leftarrow \alpha_t \hat{\mathbf{x}}_0^\prime+\beta_t\hat{\mathbf{x}}_1^\prime +\sqrt{t(1-t)\gamma^2}\mathbf{z}$ \hfill $\% \ \text{DBI}\,~~$
    \STATE $\boldsymbol{\theta}^{\star}\leftarrow\mathrm{Adam}(\boldsymbol{\theta},\nabla_{\boldsymbol{\theta}}\mathcal{L}_3(\boldsymbol{\theta}))$
    \STATE $\log r(\mathbf{x})\leftarrow \mathsf{odeint\_adjoint}(s_{\boldsymbol{\theta}^{\star}}^{t},(0,1),\mathbf{x})$
    \ENSURE estimated log density ratio $\log r(\mathbf{x})$.
\end{algorithmic}
\end{algorithm}

% \vspace{-5mm}

% \begin{proposition}[Smoothness of DSBI Density Ratio]
%     Under the DSBI interpolant $X_t = \alpha_t X_0 + \beta_t X_1 + \sqrt{t(1-t)\gamma^2} Z_t$ with $(X_0,X_1) \sim \pi_{2\gamma^2}^\star$ (OT-optimal coupling), the time-dependent density ratio $r_t(x) = q_t(x)/q_0(x)$ satisfies:
% \begin{equation}
% \|\nabla \log r_t(x)\| \leq \frac{C}{\gamma \sqrt{t(1-t)}}, 
% \end{equation}
% where $C$ depends on $\|q_0\|_{C^2}, \|q_1\|_{C^2}$. For DDBI (with independent coupling), the bound relaxes to $C/\sqrt{t(1-t)}$.
% \end{proposition}

% \begin{theorem}
%     Let $q_0, q_1 \in \mathcal{P}_2(\mathbb{R}^d)$ be distributions with finite second moments, and $\epsilon = \Theta(\gamma^2)$ the dequantization noise. Define the variance-to-transport ratio as:
%     \begin{equation}
%     \kappa := \frac{\mathrm{Var}(X_0) + \mathrm{Var}(X_1)}{W_2^2(q_0, q_1)}.
%     \end{equation}

% Suppose $\kappa > 1$ (i.e., the sum of variances dominates the squared Wasserstein distance) and $\gamma^2 \ll \min(1, W_2^2(q_0, q_1)/d)$. Then, there exists a critical value:
% \begin{equation}
% \gamma_{\max} = \sqrt{ \frac{W_2^2(q_0, q_1)}{\mathrm{Var}(X_0) + \mathrm{Var}(X_1) - W_2^2(q_0, q_1)} },
% \end{equation}
% such that for all $\gamma \in (0, \gamma_{\max})$, the interpolation errors satisfy:
% \begin{equation}
% \mathcal{E}_{\text{DSBI}}  < \mathcal{E}_{\text{DDBI}}.
% \end{equation}
% \end{theorem}

\section{Experiments}
For experiments involving $\text{D}^3\text{RE}$, we implement both DDBI and DSBI. %  \footnote{Code is available at \href{https://github.com/Hoemr/Dequantified-Diffusion-Bridge-Density-Ratio-Estimation.git}{https://github.com/Hoemr/Dequantified-Diffusion-Bridge-Density-Ratio-Estimation.git}.}
Unless specified otherwise, we use the following settings: $\alpha_t=1-t, \beta_t=t,\gamma^2=0.5,\varepsilon=1e-5$ and $\lambda(t)=\gamma^2t(1-t)$. 
Under this configuration, the interpolant $\mathbf{I}(\mathbf{X}_0,\mathbf{X}_1, t)=(1-t)\mathbf{X}_0+t\mathbf{X}_1$ aligns with the Benamou-Brenier solution to the optimal transport problem in Euclidean space \citep{mccann1997convexity}. 
The parameterized score model is trained with time score matching loss $\mathcal{L}_3$ and optimized with Adam optimization method.

\subsection{Density Estimation}
% \vspace{-2mm}
Let $r(\mathbf{x})=\frac{q_1(\mathbf{x})}{q_0(\mathbf{x})}$ be the target density ratio, where $q_1(\mathbf{x})$ is an intractable data  distribution, 
and $q_0(\mathbf{x})$ is the simpler, tractable noise distribution.
Once the estimated density ratio $r_{\boldsymbol{\theta^{\star}}}$ is obtained, the log-density of $q_1$ can be approximated as $\log q_1(\mathbf{x})\approx\log r_{\boldsymbol{\theta^{\star}}}(\mathbf{x})+\log q_0(\mathbf{x})$. 

\begin{figure}[!ht]
    \centering
    \includegraphics[width=0.99\linewidth]{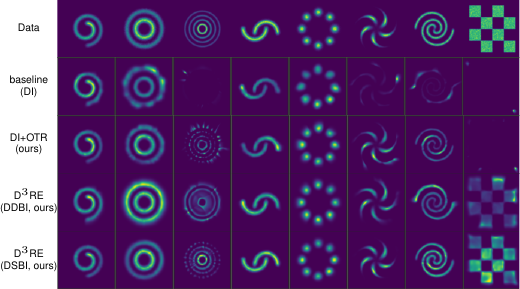}
    \caption{Density estimation results on eight 2-D synthetic datasets for different methods. 
    $\text{D}^3\text{RE}$ effectively estimates the density for both multi-modal and discontinuous distributions.%, outperforming DRE-$\infty$.
    }
    \label{fig:likelihood-estimation-toy2d-method}
\end{figure}

\textbf{2-D Synthetic Datasets.} We trained DRE-$\infty$ (baseline) and $\text{D}^3\text{RE}$ (ours) on eight 2-D synthetic datasets, including $\mathsf{swissroll}$, $\mathsf{circles}$, $\mathsf{rings}$, $\mathsf{moons}$, $\mathsf{8gaussians}$, $\mathsf{pinwheel}$, $\mathsf{2spirals}$, and $\mathsf{checkerboard}$, for 20,000 epochs using the joint score matching loss (details in \cref{appendix:joint-score-matching}). 
The density estimation results are shown in \cref{fig:likelihood-estimation-toy2d-method}. 
% \vspace{-3mm}

% \vspace{-5mm}

Our experiments demonstrate that $\text{D}^3\text{RE}$ effectively estimates the density for both multi-modal and discontinuous distributions, outperforming DRE-$\infty$. The baseline struggles, especially with complex datasets like $\mathsf{rings}$ and $\mathsf{checkerboard}$, where significant distortions occur. While the OTR trick reduces errors, it remains insufficient for intricate datasets like $\mathsf{2spirals}$ and $\mathsf{pinwheel}$. 

In contrast, the $\text{D}^3\text{RE}$ framework, leveraging both DDBI and DSBI, achieves significantly improved density estimation. DDBI captures fine-grained details, while DSBI provides robust performance across all datasets, particularly excelling in modeling complex distributions and mitigating estimation artifacts. Further comparisons across training epochs are available in \cref{appendix:likelihood-estimation-trajectory-2d-methods}.

\textbf{Energy-based Modeling on MNIST.} 
\label{section:likelihood-estimation}
We applied the proposed $\text{D}^3\text{RE}$ framework for density estimation on the MNIST dataset, leveraging pre-trained energy-based models (EBMs) \citep{choi2022density}. 
Experimental details are in \cref{appendix:likelihood-estimation-MNIST}. The results are reported in bits-per-dimension (BPD). 
Results in \cref{tab:bpd-comparison-main} show that $\text{D}^3\text{RE}$ achieves the lowest BPD values across all noise types (Gaussian, Copula, and RQ-NSF), outperforming baselines and existing methods. See also \cref{tab:bpd-comparison} in the Appendix.

Specifically, $\text{D}^3\text{RE}$ consistently surpasses DRE-$\infty$ and its variant with OTR. The DSBI method delivers the best overall results, achieving BPD values of 1.293 (Gaussian), 1.170 (Copula), and 1.066 (RQ-NSF), demonstrating its robustness and effectiveness in optimizing density estimates. 
Compared to traditional methods like NCE and TRE, $\text{D}^3\text{RE}$ shows significant improvements, especially under challenging noise distributions like Gaussian and Copula, where baseline methods yield higher BPD. These findings underscore $\text{D}^3\text{RE}$'s superior performance in accurately estimating densities and modeling complex data distributions.

% \vspace{-6mm}

\begin{table}[htbp]
\centering
\caption{Comparison of the estimated densities on MNIST dataset based on pre-trained energy-based models. The results are reported in bits-per-dim (BPD). Lower is better. The reported results for NCE and TRE are from  \citet{rhodes2020telescoping}.}
 \vskip 0.15in
\begin{tabular}{cccc}
\toprule
\textbf{Method} & \textbf{Noise type} & \textbf{Noise} & \textbf{BPD} ($\downarrow$)  \\
\midrule
NCE & Gaussian & 2.01 & 1.96 \\
TRE & Gaussian & 2.01 & 1.39  \\
DRE-$\infty$  & Gaussian & 2.01 & 1.33 \\
\midrule
\cellcolor{gray!10}DRE-$\infty$+OTR & \cellcolor{gray!10}Gaussian & \cellcolor{gray!10}2.01 & \cellcolor{gray!10}1.313  \\
\cellcolor{gray!10}$\text{D}^3\text{RE}$ (DDBI) & \cellcolor{gray!10}Gaussian & \cellcolor{gray!10}2.01 & \cellcolor{gray!10}1.297 \\
\cellcolor{gray!10}$\text{D}^3\text{RE}$ (DSBI) & \cellcolor{gray!10}Gaussian & \cellcolor{gray!10}2.01 & \cellcolor{gray!10}\textbf{1.293}\\
\midrule
NCE  & Copula & 1.40 & 1.33 \\
TRE & Copula & 1.40 & 1.24  \\
DRE-$\infty$  & Copula & 1.40 & 1.21  \\
\midrule
\cellcolor{gray!10}DRE-$\infty$+OTR & \cellcolor{gray!10}Copula & \cellcolor{gray!10}1.40 & \cellcolor{gray!10}1.204  \\
\cellcolor{gray!10}$\text{D}^3\text{RE}$ (DDBI) & \cellcolor{gray!10}Copula & \cellcolor{gray!10}1.40 & \cellcolor{gray!10}1.193 \\
\cellcolor{gray!10}$\text{D}^3\text{RE}$ (DSBI) & \cellcolor{gray!10}Copula & \cellcolor{gray!10}1.40 & \cellcolor{gray!10}\textbf{1.170} \\
\midrule
NCE  & RQ-NSF & 1.12 & 1.09  \\
TRE  & RQ-NSF & 1.12 & 1.09 \\
DRE-$\infty$  & RQ-NSF & 1.12 & 1.09 \\
\midrule
\cellcolor{gray!10}DRE-$\infty$+OTR & \cellcolor{gray!10}RQ-NSF & \cellcolor{gray!10}1.12 & \cellcolor{gray!10}1.072  \\
\cellcolor{gray!10}$\text{D}^3\text{RE}$ (DDBI) & \cellcolor{gray!10}RQ-NSF & \cellcolor{gray!10}1.12 & \cellcolor{gray!10}1.072  \\
\cellcolor{gray!10}$\text{D}^3\text{RE}$ (DSBI) & \cellcolor{gray!10}RQ-NSF & \cellcolor{gray!10}1.12 & \cellcolor{gray!10}\textbf{1.066} \\
\bottomrule
\end{tabular}
% \vskip -0.1in
\label{tab:bpd-comparison-main}
\end{table}
% \vspace{-3mm}

\begin{figure*}[t]
    \centering
    \subfigure[$d=40$]{\includegraphics[width=0.32\linewidth]{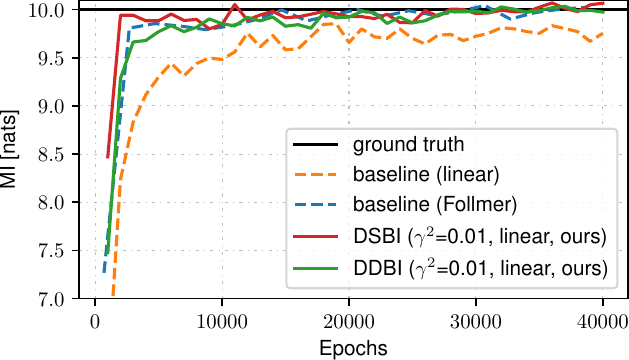}\label{fig:MI-40d}}
    \hfill
    \subfigure[$d=80$]{\includegraphics[width=0.32\linewidth]{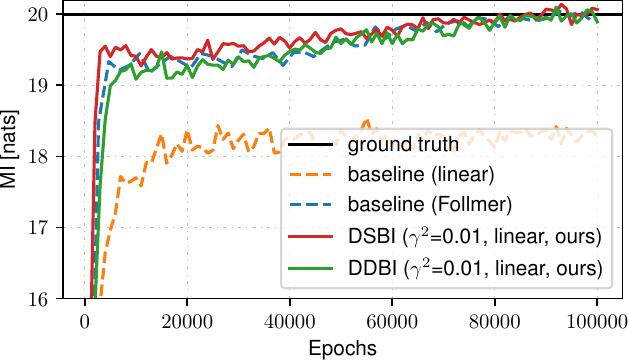}\label{fig:MI-80d}}
    \hfill
    \subfigure[$d=120$]{\includegraphics[width=0.32\linewidth]{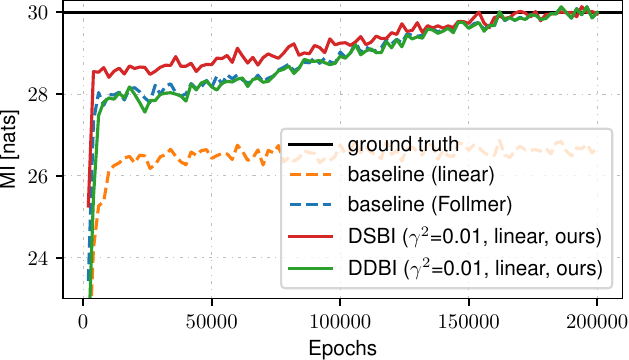}\label{fig:MI-120d}}
    \hfill
    \caption{Evolution of estimated MI across epochs with varying methods and dimensions $d = \{40, 80, 120\}$.  
    $\text{D}^3\text{RE}$ outperforms the baseline in both speed and precision.
    DRE-$\infty$ \citep{choi2022density} is regarded as the `baseline' in this experiment.
    }
    \label{fig:MI-evolution}
\end{figure*}

\subsection{Mutual Information Estimation}
% \vspace{-2mm}
Mutual information (MI) measures the dependency  between two random variables $\mathbf{X}\sim p(\mathbf{x})$ and $\mathbf{Y}\sim q(\mathbf{y})$, defined as $\text{MI}(\mathbf{X},\mathbf{Y})=\mathbb{E}_{ p(\mathbf{x},\mathbf{y})}\left[\log\frac{p(\mathbf{x},\mathbf{y})}{p(\mathbf{x})q(\mathbf{y})}\right]$. 
Here $p(\mathbf{x},\mathbf{y})$ be their joint density.
$q(\mathbf{y})=\mathcal{N}(\mathbf{0},\sigma^2\mathbf{E}_d)$ and $p(\mathbf{x})=\mathcal{N}(\mathbf{0},\mathbf{E}_d)$, with $\sigma^2=1e-6$ and $d=\{40, 80, 120 \}$, are two $d$-dimensional correlated Gaussian distributions.
The experimental setup in  DRE-$\infty$ \citep{choi2022density} is adapted to implement $\text{D}^3\text{RE}$.
DRE-$\infty$ serves as the benchmark.
More details can be found in  \cref{appendix:mutual-information}.

The evolution of estimated MI across epochs for $d=\{40,80,120\}$, comparing  $\text{D}^3\text{RE}$ with DRE-$\infty$, are analyzed. 
Results in \cref{fig:MI-evolution} show that the red (DSBI) and green (DDBI) curves outperform the blue and yellow (DRE-$\infty$) curves in two aspects. 
First, $\text{D}^3\text{RE}$ converges to the true MI value more rapidly as it expands trajectory sets (see \cref{corollary:path-set-expansion}), improving interpolation accuracy. Second, it exhibits greater stability with fewer fluctuations around the true MI, indicating more reliable estimates.

We conclude that $\text{D}^3\text{RE}$ outperforms the baseline in both speed and precision. For $d=120$, the MI estimated by $\text{D}^3\text{RE}$ is much robust than that of DRE-$\infty$.

% \vspace{-3mm}

\subsection{Analysis and Discussion}

% \vspace{-2mm}

\begin{figure}[t]
    \centering
    \includegraphics[width=0.99\linewidth]{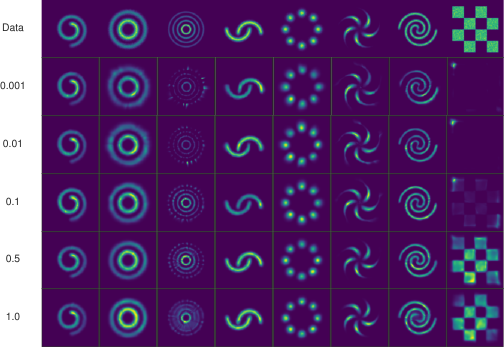}
    % \vspace{-2mm}
    \caption{Ablation study on the effect of $\gamma^2$ for density estimation on 2-D toy data. The first row displays the results for the ground truth data. Each subsequent row, from top to bottom, corresponds to $\gamma^2$ values of 0.001, 0.01, 0.1, 0.5, and 1.0, respectively.}
    \label{fig:ablation-study-all-toy}
\end{figure}

\textbf{Ablation Study on $\gamma^2$.}
The ablation study on $\gamma^2$ for density estimation (\cref{fig:ablation-study-all-toy}) reveals systematic trade-offs in performance across regularization strengths. For small $\gamma^2 = 0.001$, the model achieves rapid initial alignment with the ground truth distribution (first row) but exhibits overfitting artifacts in later epochs, manifesting as irregular density peaks and deviations from the smooth ground truth structure. Intermediate values ($\gamma^2 = 0.01$–$0.1$) demonstrate balanced behavior: $\gamma^2 = 0.01$ preserves finer details while maintaining stability, and $\gamma^2 = 0.1$ produces smoother approximations with minimal divergence from the true distribution. Larger $\gamma^2$ values ($\geq 0.5$) induce excessive regularization, leading to oversmoothed estimates that fail to capture critical modes of the 2-D data, particularly in high-density regions. Notably, $\gamma^2 = 0.1$ achieves the closest visual and structural resemblance to the ground truth, suggesting its suitability for low-dimensional tasks requiring both fidelity and robustness. These results underscore the necessity of tuning $\gamma^2$ to mitigate under-regularization artifacts while preserving distributional complexity.
% \vspace{-2mm}

% \vspace{-5mm}

\begin{figure*}[t]
    \centering
    \subfigure[$d=40$]{\includegraphics[width=0.32\linewidth]{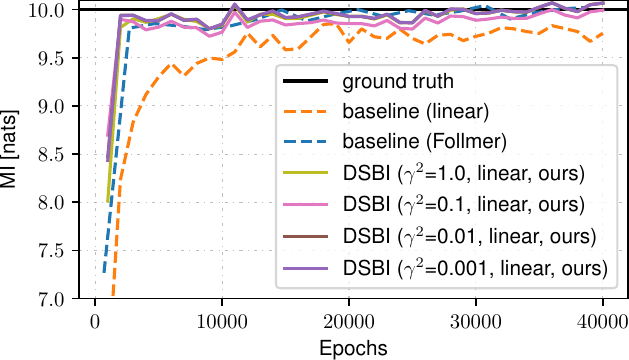}\label{fig:MI-40d-gamma}}
    \hfill
    \subfigure[$d=80$]{\includegraphics[width=0.32\linewidth]{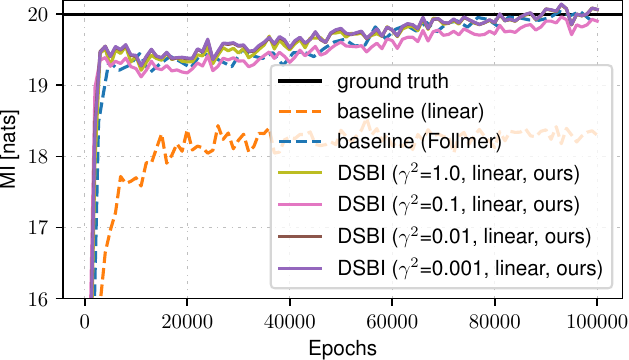}\label{fig:MI-80d-gamma}}
    \hfill
    \subfigure[$d=120$]{\includegraphics[width=0.32\linewidth]{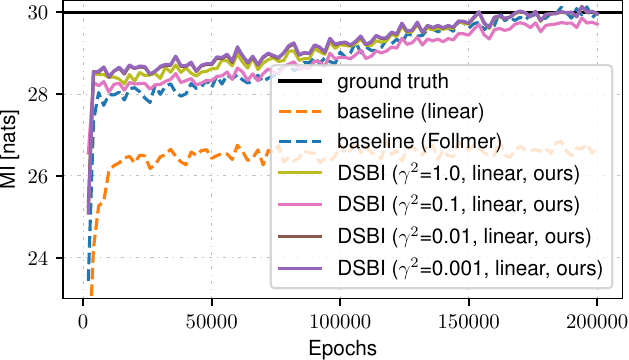}\label{fig:MI-120d-gamma}}
    \hfill
    \caption{Evolution of estimated MI across epochs with varying $\gamma^2=\{0.001,0.01,0.1,1.0\}$ and dimensions $d = \{40, 80, 120\}$. For all dimensions ($d = \{40, 80, 120\}$), smaller $\gamma^2$ values ($\leq 0.01$) lead to faster convergence. 
    % More results can be found in \cref{fig:MI-evolution-gamma}.
    }
    \label{fig:MI-evolution-gamma-main}
\end{figure*}

The ablation study on varying $\gamma^2$ values (\cref{fig:MI-evolution-gamma-main}) shows distinct convergence behaviors in MI estimation. For all dimensions ($d = \{40, 80, 120\}$), smaller $\gamma^2$ values ($\leq 0.01$) lead to faster convergence, especially in lower dimensions ($d = 40$), but excessively small values ($\gamma^2 = 0.001$) cause instability later. Larger $\gamma^2$ values ($\geq 0.1$) converge more slowly but stabilize over time, particularly in higher dimensions ($d = 120$). $\gamma^2 = 0.1$ offers a balance between speed and stability across all dimensions, suggesting that moderate regularization provides the best MI estimation performance.
More results are provided in \cref{appendix:ablation-study-gamma}.

% \vspace{-2mm}

\begin{figure}[t]
	\centering
	\includegraphics[width=0.99\linewidth]{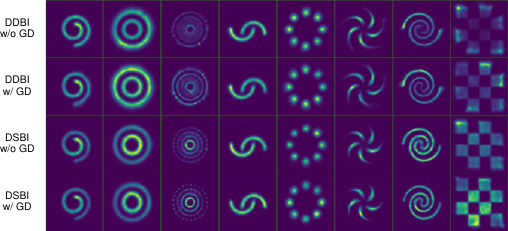}
	\caption{Ablation study on GD for eight 2-D synthetic datasets.}
	\label{fig:ablation_gd}
\end{figure}

\textbf{Ablation Study on GD.}
To evaluate the effectiveness of the proposed GD module, we conduct an ablation study by comparing model performance with and without GD, as shown in \cref{fig:ablation_gd}.
Visually, both DDBI and DSBI show clear improvements in density estimation when GD is applied. Without GD, the estimated densities appear blurrier and miss fine structural details, whereas incorporating GD yields sharper and more realistic patterns.
% \vspace{-3mm}

% \vspace{-5mm}

\begin{figure}[ht]
	\centering
	\includegraphics[width=0.99\linewidth]{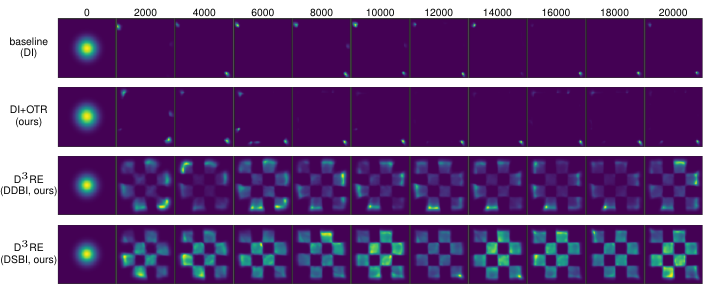}
	\caption{Density estimation results on $\mathsf{checkerboard}$ for different methods during training (see more results in \cref{fig:likelihood-estimation-trajectory-2d-method-swissroll,fig:likelihood-estimation-trajectory-2d-method-circles,fig:likelihood-estimation-trajectory-2d-method-rings,fig:likelihood-estimation-trajectory-2d-method-moons,fig:likelihood-estimation-trajectory-2d-method-8gaussians,fig:likelihood-estimation-trajectory-2d-method-pinwheel,fig:likelihood-estimation-trajectory-2d-method-2spirals,fig:likelihood-estimation-trajectory-2d-method-checkerboard}).}
	\label{fig:likelihood-estimation-trajectory-2d-method-checkerboard-main}
\end{figure}

\textbf{Ablation Study on OTR.}
We conduct an ablation study to evaluate the role of OTR, comparing models without OTR (baseline, DI), models with OTR (DI+OTR), and models from the $\text{D}^3\text{RE}$ framework (DDBI and DSBI). 
In \cref{fig:likelihood-estimation-toy2d-method,fig:likelihood-estimation-trajectory-2d-method-checkerboard-main}, DI generates distorted and misaligned intermediate distributions. This shows its limited ability to align with the target distribution. 
DI+OTR improves alignment but remains suboptimal. 
Models from the $\text{D}^3\text{RE}$ further enhance distribution quality, with DSBI achieving the most precise alignment. This underscores OTR’s crucial role in improving intermediate distributions.
% \vspace{-3mm}
% \vspace{-5mm}

\cref{fig:MI-evolution} compares MI estimation for DDBI and DSBI across dimensions ($d=80, 120$). Both outperform baseline methods, but DSBI converges faster and remains closer to the ground truth. The advantage of OTR becomes more pronounced in high dimensions ($d=120$), where DSBI significantly outperforms DDBI in both speed and accuracy.
% % \vspace{-5mm}

Overall, OTR improves intermediate distribution alignment. When combined with diffusion bridges and GD, as in DSBI, it enables more accurate density estimation. Further comparisons are presented in \cref{appendix:ablation-study-OTR}.

\begin{figure}[!b]
    \centering
    \includegraphics[width=0.99\linewidth]{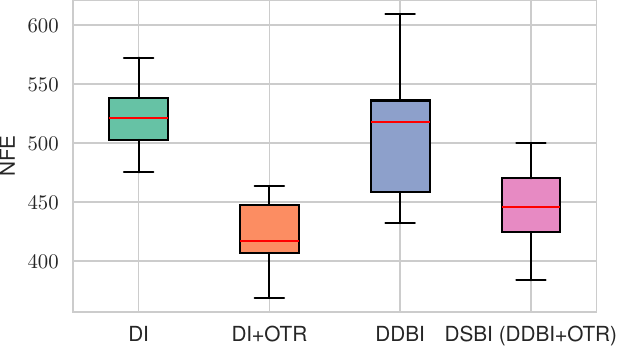}
    \caption{Comparison of NFE for four methods in density ratio estimation task. Applying OTR significantly reduces NFE.}
    \label{fig:nfe-comparison-density-ratio-toy2toy}
\end{figure}

\textbf{Number of Function Evaluations. }
We analyze the impact of OTR on NFE, noting that DI and DDBI do not utilize OTR. 
It shows that applying OTR significantly reduces NFE. \cref{fig:nfe-comparison-density-ratio-toy2toy} compares NFE across four methods in DRE.
% % \vspace{-2mm}
% % \vspace{-3mm}
The first approach exhibits the highest NFE, indicating reliance on iterative procedures requiring repeated function evaluations. The second approach achieves a moderate reduction in NFE, likely by minimizing redundant evaluations through minimized transport costs.

\section{Related Works}
\textbf{Density Ratio Estimation.} DRE is an essential technique in machine learning but faces challenges in high-dimensional settings and when distributions are significantly different. Early methods \citep{sugiyama2012density, gutmann2012noise} often struggled with instability, known as the density-chasm problem, in these scenarios. To overcome these challenges, TRE \citep{rhodes2020telescoping}, an extension of NCE \citep{gutmann2012noise}, introduced a divide-and-conquer approach, breaking the problem into simpler subproblems for better performance. DRE-$\infty$ \citep{choi2022density} further advanced this by interpolating between distributions through an infinite series of bridge distributions, improving stability and accuracy. F-DRE \citep{choi2021featurized} used an invertible generative model to map distributions into a common feature space before estimation. Recent methods, such as \citet{kato2021non}, have addressed overfitting in flexible models, and \citet{nagumodensity,luo2024game} focused on improving robustness to outliers. MDRE \citep{srivastava2023estimating} tackled distribution shift through multi-class classification, offering an alternative to binary classification in high-discrepancy cases. Additionally, geometric approaches, like \citet{kimura2024density}, have enhanced DRE accuracy by incorporating the geometry of statistical manifolds. Building on these advancements, our work proposes a novel method to improve both the accuracy and robustness of high-dimensional DRE.

\textbf{Diffusion Bridge.} Denoising diffusion implicit models (DDIMs) \citep{song2020denoising} have been proposed as an efficient alternative to denoising diffusion probabilistic models (DDPMs) \citep{ho2020denoising}, which require simulating a Markov chain for many steps to generate samples. 
Diffusive interpolants \citep{albergo2023stochastic} provide a unifying framework of flow-based models and diffusion models, bridging arbitrary distributions using continuous-time stochastic processes.
DDBMs \citep{zhou2023denoising} are  proposed  as a natural alternative to cumbersome methods like guidance or projected sampling in generative processes. 
Our proposed DDBI and DSBI build upon diffusive  interpolants and DDBMs by incorporating  Brownian bridge into the interpolation strategy construction. 

\section{Conclusions}
In this work, we propose $\text{D}^3\text{RE}$, a unified, robust and efficient framework for density ratio estimation. It provides the first framework for directly addressing both the density-chasm and support-chasm problems, enabling uniformly approximated density-ratio estimation (\cref{proposition:uniform-estimation-formula}). By incorporating diffusion bridges and GD, we construct DDBI, which expands support coverage (\cref{theorem:support-set-expansion}, \cref{corollary:path-set-expansion}) and stabilizes the time score near boundaries (\cref{corollary:upper-bound-time-score-DDBI-DSBI}).
Building upon DDBI,  OTR is incorporated to derive the DSBI, which offers more efficient and stable density ratio estimation (\cref{theorem:upper-bound-of-DDBI-DSBI}) by solving the Schr{\"o}dinger bridge problem (\cref{proposition:solution-to-SB-problem}).
Together, DDBI and DSBI form the core of the $\text{D}^3\text{RE}$ framework, enabling uniformly approximated and stable density-ratio estimation (\cref{proposition:uniform-estimation-formula}).
Extensive experiments validate these findings, demonstrating the superior performance of $\text{D}^3\text{RE}$ in tasks such as density-ratio estimation on synthetic data, mutual information estimation, and density estimation. 

While $\text{D}^3\text{RE}$ advances density ratio estimation methods, several directions remain open. Future work could explore adaptive or learned solvers to reduce function evaluation overhead, as well as more expressive interpolants to further improve robustness in handling complex or multi-modal distributions.

\section*{Acknowledgements}
This research work is supported by the Fundamental Research Program of Guangdong, China, under Grant  2023A1515011281.

% \textbf{Do not} include acknowledgements in the initial version of
% the paper submitted for blind review.

% If a paper is accepted, the final camera-ready version can (and
% usually should) include acknowledgements.  Such acknowledgements
% should be placed at the end of the section, in an unnumbered section
% that does not count towards the paper page limit. Typically, this will 
% include thanks to reviewers who gave useful comments, to colleagues 
% who contributed to the ideas, and to funding agencies and corporate 
% sponsors that provided financial support.

\section*{Impact Statement}
This paper presents work whose goal is to advance the field of Machine Learning. 
While direct societal impacts are limited, future extensions to applied domains (e.g., via our open-source codebase) should incorporate domain-specific ethical reviews per deployment contexts.


\bibliography{DSBI}

\begin{thebibliography}{40}
\providecommand{\natexlab}[1]{#1}
\providecommand{\url}[1]{\texttt{#1}}
\expandafter\ifx\csname urlstyle\endcsname\relax
  \providecommand{\doi}[1]{doi: #1}\else
  \providecommand{\doi}{doi: \begingroup \urlstyle{rm}\Url}\fi

\bibitem[Albergo et~al.(2023)Albergo, Boffi, and
  Vanden-Eijnden]{albergo2023stochastic}
Albergo, M.~S., Boffi, N.~M., and Vanden-Eijnden, E.
\newblock Stochastic interpolants: A unifying framework for flows and
  diffusions.
\newblock \emph{arXiv preprint arXiv:2303.08797}, 2023.

\bibitem[Burda et~al.(2015)Burda, Grosse, and Salakhutdinov]{burda2015accurate}
Burda, Y., Grosse, R., and Salakhutdinov, R.
\newblock Accurate and conservative estimates of mrf log-likelihood using
  reverse annealing.
\newblock In \emph{Artificial Intelligence and Statistics}, 2015.

\bibitem[Choi et~al.(2021)Choi, Liao, and Ermon]{choi2021featurized}
Choi, K., Liao, M., and Ermon, S.
\newblock Featurized density ratio estimation.
\newblock In \emph{Uncertainty in Artificial Intelligence}, 2021.

\bibitem[Choi et~al.(2022)Choi, Meng, Song, and Ermon]{choi2022density}
Choi, K., Meng, C., Song, Y., and Ermon, S.
\newblock Density ratio estimation via infinitesimal classification.
\newblock In \emph{Artificial Intelligence and Statistics}, 2022.

\bibitem[Colombo et~al.(2021)Colombo, Piantanida, and Clavel]{colombo2021novel}
Colombo, P., Piantanida, P., and Clavel, C.
\newblock A novel estimator of mutual information for learning to disentangle
  textual representations.
\newblock In \emph{Annual Meeting of the Association for Computational
  Linguistics}, 2021.

\bibitem[Cuturi(2013)]{cuturi2013sinkhorn}
Cuturi, M.
\newblock Sinkhorn distances: Lightspeed computation of optimal transport.
\newblock In \emph{Advances in Neural Information Processing Systems}, 2013.

\bibitem[De~Bortoli et~al.(2021)De~Bortoli, Thornton, Heng, and
  Doucet]{de2021diffusion}
De~Bortoli, V., Thornton, J., Heng, J., and Doucet, A.
\newblock Diffusion {S}chr{\"o}dinger bridge with applications to score-based
  generative modeling.
\newblock In \emph{Advances in Neural Information Processing Systems}, 2021.

\bibitem[Doob \& Doob(1984)Doob and Doob]{doob1984classical}
Doob, J.~L. and Doob, J.
\newblock \emph{Classical Potential Theory and its Probabilistic Counterpart},
  volume 262.
\newblock Springer, 1984.

\bibitem[Durkan et~al.(2019)Durkan, Bekasov, Murray, and
  Papamakarios]{durkan2019neural}
Durkan, C., Bekasov, A., Murray, I., and Papamakarios, G.
\newblock Neural spline flows.
\newblock In \emph{Advances in Neural Information Processing Systems}, 2019.

\bibitem[Gutmann \& Hyv{\"a}rinen(2012)Gutmann and
  Hyv{\"a}rinen]{gutmann2012noise}
Gutmann, M.~U. and Hyv{\"a}rinen, A.
\newblock Noise-contrastive estimation of unnormalized statistical models, with
  applications to natural image statistics.
\newblock \emph{Journal of Machine Learning Research}, 13\penalty0 (2), 2012.

\bibitem[Ho et~al.(2020)Ho, Jain, and Abbeel]{ho2020denoising}
Ho, J., Jain, A., and Abbeel, P.
\newblock Denoising diffusion probabilistic models.
\newblock In \emph{Advances in Neural Information Processing Systems}, 2020.

\bibitem[Kato \& Teshima(2021)Kato and Teshima]{kato2021non}
Kato, M. and Teshima, T.
\newblock Non-negative {B}regman divergence minimization for deep direct
  density ratio estimation.
\newblock In \emph{International Conference on Machine Learning}, 2021.

\bibitem[Kimura \& Bondell(2025)Kimura and Bondell]{kimura2024density}
Kimura, M. and Bondell, H.
\newblock Density ratio estimation via sampling along generalized geodesics on
  statistical manifolds.
\newblock In \emph{Artificial Intelligence and Statistics}, 2025.

\bibitem[L{\'e}onard(2014)]{leonard2014some}
L{\'e}onard, C.
\newblock Some properties of path measures.
\newblock \emph{S{\'e}minaire de Probabilit{\'e}s XLVI}, pp.\  207--230, 2014.

\bibitem[Li et~al.(2024{\natexlab{a}})Li, Chen, Liu, Yang, Zeng, and
  Zhou]{li2024neural}
Li, J., Chen, W., Liu, Y., Yang, J., Zeng, D., and Zhou, Z.
\newblock Neural ordinary differential equation networks for fintech
  applications using internet of things.
\newblock \emph{IEEE Internet of Things Journal}, 2024{\natexlab{a}}.

\bibitem[Li et~al.(2024{\natexlab{b}})Li, Chen, Zhou, Yang, and
  Zeng]{li2024deepar}
Li, J., Chen, W., Zhou, Z., Yang, J., and Zeng, D.
\newblock Deepar-attention probabilistic prediction for stock price series.
\newblock \emph{Neural Computing and Applications}, 36\penalty0 (25):\penalty0
  15389--15406, 2024{\natexlab{b}}.

\bibitem[Li et~al.(2023)Li, Chen, and Zeng]{li2023scire}
Li, S., Chen, W., and Zeng, D.
\newblock Scire-solver: Accelerating diffusion models sampling by
  score-integrand solver with recursive difference.
\newblock \emph{arXiv preprint arXiv:2308.07896}, 2023.

\bibitem[Luo et~al.(2024)Luo, Bao, Zhou, and Dang]{luo2024game}
Luo, R., Bao, J., Zhou, Z., and Dang, C.
\newblock Game-theoretic defenses for robust conformal prediction against
  adversarial attacks in medical imaging.
\newblock \emph{arXiv preprint arXiv:2411.04376}, 2024.

\bibitem[McCann(1997)]{mccann1997convexity}
McCann, R.~J.
\newblock A convexity principle for interacting gases.
\newblock \emph{Advances in Mathematics}, 128\penalty0 (1):\penalty0 153--179,
  1997.

\bibitem[Nagumo \& Fujisawa(2024)Nagumo and Fujisawa]{nagumodensity}
Nagumo, R. and Fujisawa, H.
\newblock Density ratio estimation with doubly strong robustness.
\newblock In \emph{International Conference on Machine Learning}, 2024.

\bibitem[Neal(2001)]{neal2001annealed}
Neal, R.~M.
\newblock Annealed importance sampling.
\newblock \emph{Statistics and Computing}, 11:\penalty0 125--139, 2001.

\bibitem[Pooladian et~al.(2023)Pooladian, Ben-Hamu, Domingo-Enrich, Amos,
  Lipman, and Chen]{pooladian2023multisample}
Pooladian, A.-A., Ben-Hamu, H., Domingo-Enrich, C., Amos, B., Lipman, Y., and
  Chen, R.~T.
\newblock Multisample flow matching: Straightening flows with minibatch
  couplings.
\newblock In \emph{International Conference on Machine Learning}, 2023.

\bibitem[Rhodes et~al.(2020)Rhodes, Xu, and Gutmann]{rhodes2020telescoping}
Rhodes, B., Xu, K., and Gutmann, M.~U.
\newblock Telescoping density-ratio estimation.
\newblock In \emph{Advances in Neural Information Processing Systems}, 2020.

\bibitem[Schr{\"o}dinger(1932)]{schrodinger1932theorie}
Schr{\"o}dinger, E.
\newblock Sur la th{\'e}orie relativiste de l'{\'e}lectron et
  l'interpr{\'e}tation de la m{\'e}canique quantique.
\newblock In \emph{Annales de l'institut Henri Poincar{\'e}}, volume~3, pp.\
  269--310, 1932.

\bibitem[Song et~al.(2020)Song, Meng, and Ermon]{song2020denoising}
Song, J., Meng, C., and Ermon, S.
\newblock Denoising diffusion implicit models.
\newblock In \emph{International Conference on Learning Representations}, 2020.

\bibitem[Song \& Ermon(2020)Song and Ermon]{song2020improved}
Song, Y. and Ermon, S.
\newblock Improved techniques for training score-based generative models.
\newblock In \emph{Advances in Neural Information Processing Systems}, 2020.

\bibitem[Song et~al.(2021)Song, Sohl-Dickstein, Kingma, Kumar, Ermon, and
  Poole]{song2020score}
Song, Y., Sohl-Dickstein, J., Kingma, D.~P., Kumar, A., Ermon, S., and Poole,
  B.
\newblock Score-based generative modeling through stochastic differential
  equations.
\newblock In \emph{International Conference on Learning Representations}, 2021.

\bibitem[Srivastava et~al.(2023)Srivastava, Han, Xu, Rhodes, and
  Gutmann]{srivastava2023estimating}
Srivastava, A., Han, S., Xu, K., Rhodes, B., and Gutmann, M.~U.
\newblock Estimating the density ratio between distributions with high
  discrepancy using multinomial logistic regression.
\newblock \emph{arXiv preprint arXiv:2305.00869}, 2023.

\bibitem[Sugiyama et~al.(2012)Sugiyama, Suzuki, and
  Kanamori]{sugiyama2012density}
Sugiyama, M., Suzuki, T., and Kanamori, T.
\newblock Density-ratio matching under the bregman divergence: a unified
  framework of density-ratio estimation.
\newblock \emph{Annals of the Institute of Statistical Mathematics},
  64:\penalty0 1009--1044, 2012.

\bibitem[Thomas et~al.(2022)Thomas, Dutta, Corander, Kaski, and
  Gutmann]{thomas2022likelihood}
Thomas, O., Dutta, R., Corander, J., Kaski, S., and Gutmann, M.~U.
\newblock Likelihood-free inference by ratio estimation.
\newblock \emph{Bayesian Analysis}, 17\penalty0 (1):\penalty0 1--31, 2022.

\bibitem[Tong et~al.(2024)Tong, FATRAS, Malkin, Huguet, Zhang, Rector-Brooks,
  Wolf, and Bengio]{tong2023improving}
Tong, A., FATRAS, K., Malkin, N., Huguet, G., Zhang, Y., Rector-Brooks, J.,
  Wolf, G., and Bengio, Y.
\newblock Improving and generalizing flow-based generative models with
  minibatch optimal transport.
\newblock \emph{Transactions on Machine Learning Research}, 2024.

\bibitem[Villani et~al.(2009)]{villani2009optimal}
Villani, C. et~al.
\newblock \emph{Optimal Transport: Old and New}, volume 338.
\newblock Springer, 2009.

\bibitem[Xin et~al.(2024)]{xin2024v}
Xin, Y. et~al.
\newblock V-{PETL} bench: A unified visual parameter-efficient transfer
  learning benchmark.
\newblock In \emph{Advances in Neural Information Processing Systems}, 2024.

\bibitem[Xu et~al.(2024)Xu, Zeng, and Paisley]{xu2024sparse}
Xu, J., Zeng, D., and Paisley, J.
\newblock Sparse inducing points in deep {G}aussian processes: {E}nhancing
  modeling with denoising diffusion variational inference.
\newblock In \emph{International Conference on Machine Learning}, 2024.

\bibitem[Zhao et~al.(2023)Zhao, Chen, and Wang]{zhao2023learning}
Zhao, S., Chen, W., and Wang, T.
\newblock Learning few-shot sample-set operations for noisy multi-label aspect
  category detection.
\newblock In \emph{International Joint Conference on Artificial Intelligence},
  2023.

\bibitem[Zhao et~al.(2025)Zhao, Chen, Wang, Yao, Lu, and Zheng]{zhao2025less}
Zhao, S., Chen, W., Wang, T., Yao, J., Lu, D., and Zheng, J.
\newblock Less is enough: Relation graph guided few-shot learning for
  multi-label aspect category detection.
\newblock In \emph{IEEE International Conference on Acoustics, Speech and
  Signal Processing}, 2025.

\bibitem[Zhou et~al.(2024{\natexlab{a}})Zhou, Lou, Khanna, and
  Ermon]{zhou2023denoising}
Zhou, L., Lou, A., Khanna, S., and Ermon, S.
\newblock Denoising diffusion bridge models.
\newblock In \emph{International Conference on Learning Representations},
  2024{\natexlab{a}}.

\bibitem[Zhou et~al.(2024{\natexlab{b}})Zhou, Ye, Wang, Jiang, Lee, Xie, and
  Zhang]{zhou2024enhancing}
Zhou, X., Ye, W., Wang, Y., Jiang, C., Lee, Z., Xie, R., and Zhang, S.
\newblock Enhancing in-context learning via implicit demonstration
  augmentation.
\newblock In \emph{Annual Meeting of the Association for Computational
  Linguistics (Volume 1: Long Papers)}, 2024{\natexlab{b}}.

\bibitem[Zhou et~al.(2025{\natexlab{a}})Zhou, Ye, Lee, Zou, and
  Zhang]{zhou2025valuing}
Zhou, X., Ye, W., Lee, Z., Zou, L., and Zhang, S.
\newblock Valuing training data via causal inference for in-context learning.
\newblock \emph{IEEE Transactions on Knowledge and Data Engineering},
  37\penalty0 (6):\penalty0 3824 -- 3840, 2025{\natexlab{a}}.

\bibitem[Zhou et~al.(2025{\natexlab{b}})Zhou, Zhang, Lee, Ye, and
  Zhang]{zhou2025hademif}
Zhou, X., Zhang, M., Lee, Z., Ye, W., and Zhang, S.
\newblock Ha{D}e{M}i{F}: Hallucination detection and mitigation in large
  language models.
\newblock In \emph{International Conference on Learning Representations},
  2025{\natexlab{b}}.

\end{thebibliography}
\bibliographystyle{icml2025}

%%%%%%%%%%%%%%%%%%%%%%%%%%%%%%%%%%%%%%%%%%%%%%%%%%%%%%%%%%%%%%%%%%%%%%%%%%%%%%%
%%%%%%%%%%%%%%%%%%%%%%%%%%%%%%%%%%%%%%%%%%%%%%%%%%%%%%%%%%%%%%%%%%%%%%%%%%%%%%%
% APPENDIX
%%%%%%%%%%%%%%%%%%%%%%%%%%%%%%%%%%%%%%%%%%%%%%%%%%%%%%%%%%%%%%%%%%%%%%%%%%%%%%%
%%%%%%%%%%%%%%%%%%%%%%%%%%%%%%%%%%%%%%%%%%%%%%%%%%%%%%%%%%%%%%%%%%%%%%%%%%%%%%%
\newpage
\appendix
\onecolumn
	
\section{Proofs}
\begin{assumption}
    \label{ass:smoothness}
    Let $ q_0, q_1: \mathbb{R}^d \to \mathbb{R}_+ $ be probability density functions satisfying: (1) $ q_0, q_1 \in C^2(\mathbb{R}^d) $, i.e., $ q_0 $ and $ q_1 $ are twice differentiable and have bounded second derivatives: $\|\nabla_{\mathbf{x}}^2 q_0\|_{L^\infty}, \|\nabla_{\mathbf{x}}^2 q_1\|_{L^\infty}< \infty$; (2) $ q_0(\mathbf{x}) > 0 $, and there exists $ c > 0 $ such that $\inf_{\mathbf{x}} q_0(\mathbf{x}) \geq c$. This condition is mild in density ratio estimation.
\end{assumption}

\begin{assumption}
\label{assumption:Lip-conditional-distri}
    The conditional distributions for given $\mathbf{X}_t=\mathbf{x}_t$, i.e.,  $q_0(\mathbf{x}\mid\mathbf{x}_t)$ and $q_1(\mathbf{x}\mid\mathbf{x}_t)$ have $L$-Lipschitz scores:
\begin{equation}
  \|\nabla_{\mathbf{x}_t} \log q_0(\mathbf{x}\mid\mathbf{x}_t)\| \leq \frac{L}{\alpha_t}, \quad \|\nabla_{\mathbf{x}_t} \log q_1(\mathbf{x}\mid\mathbf{x}_t)\| \leq \frac{L}{\beta_t}.
\end{equation}
\end{assumption}

\subsection{Proof of \cref{theorem:support-set-expansion}}
\label{proof:support-set-expansion}
\begin{proof}
    We first consider the support for DI. 
    Under $\alpha_t+\beta_t=1$, the $\mathbf{X}_t$ is the convex combination of $\mathbf{X}_0$ and $\mathbf{X}_1$ for any $t\in(0,1)$, and its corresponding support, $\mathsf{supp}(q_t)$, is the \textit{convex hull} of $\mathsf{supp}(q_0)$ and $\mathsf{supp}(q_q)$, i.e., $\mathsf{supp}(q_t)=\mathsf{conv}\left(\mathsf{supp}(q_0)\cup\mathsf{supp}(q_1)\right)$. Under $\alpha_t^2 + \beta_t^2=1$, $\mathbf{X}_t$ becomes a linear combination of $\mathbf{x}_0$ and $\mathbf{x}_1$ for any $t\in(0,1)$. For both cases, $\mathsf{supp}(q_t)$ can be formulated as
    \begin{equation}
    \begin{aligned}
        \mathsf{supp}(q_t)&=\alpha_t \mathsf{supp}(q_0)+ \beta_t\mathsf{supp}(q_1) \\
        &=\{ \alpha_t \mathbf{x}_0+\beta_t \mathbf{x}_1\mid \mathbf{x}_0\in \mathsf{supp}(q_0), \mathbf{x}_1\in \mathsf{supp}(q_1), \alpha_t+\beta_t=1 \ \text{or} \ \alpha_t^2+\beta_t^2=1 \},
    \end{aligned}  \notag
    \end{equation}
    where $\mathsf{supp}(q_0)$ and $\mathsf{supp}(q_1)$ are the supports of $q_0$ and $q_1$, respectively. 

    Next, we consider the support for DBI. 
    For given coefficients $\alpha_t$ and $\beta_t$, $\mathbf{X}_t^\prime$ can be formulated as:
    \begin{equation}
        \mathbf{X}_t^\prime = \mathbf{I}(\mathbf{X}_0, \mathbf{X}_1, t) + \sqrt{t(1-t)\gamma^2 } \mathbf{Z}_t=\mathbf{X}_t+\sqrt{t(1-t)\gamma^2 } \mathbf{Z}_t.
    \end{equation}
    The coefficient $\sqrt{t(1-t)\gamma^2}$ is deterministic for a given $t$.
    Thus the support corresponding to $\mathbf{X}_t^\prime$, denoted as $\mathsf{supp}(q_t^\prime)$, can be expressed as the Minkowski sum of the supports of $q_t$ and $\mathcal{N}(\mathbf{0},\mathbf{E}_d)$:
    \begin{equation}
        \mathsf{supp}(q_t^\prime)=\mathsf{supp}(q_t)+\mathsf{supp}(\mathcal{N}(\mathbf{0},\mathbf{E}_d))=\{ \mathbf{x}+\mathbf{z}\mid \mathbf{x}\in \mathsf{supp}(q_t), z\in \mathsf{supp}(\mathcal{N}(\mathbf{0},\mathbf{E}_d))\},   \notag
    \end{equation}
    where $\mathsf{supp}(\mathcal{N}(\mathbf{0},\mathbf{E}_d))=\mathbb{R}^d$. 
    The Minkowski sum $\mathsf{supp}(q_t)+\mathsf{supp}(\mathcal{N}(\mathbf{0},\mathbf{E}_d))$ is at least as large as $\mathsf{supp}(q_t)$, i.e., $\mathsf{supp}(q_t^\prime)\supseteq\mathsf{supp}(q_t)$. This completes the proof.
\end{proof}

\subsection{Proof of \cref{corollary:path-set-expansion}}
\label{proof:path-set-expansion}

\begin{proof}
Let the trajectory sets for DI and DBI be denoted by $\mathbb{T}=\{\{\mathbf{x}_t \}_{t\in[0,1]}; \mathbf{x}_t\in\mathsf{supp}(q_t)\}$ and $\mathbb{T}^{\prime}=\{\{\mathbf{x}_t^{\prime}\}_{t\in[0,1]}; \mathbf{x}_t^{\prime}\in\mathsf{supp}(q_t^{\prime})\}$, respectively.
Let $\{\mathbf{x}_t\}_{t\in[0,1]}$ be an arbitrary element of $ \mathbb{T}$.
From \cref{theorem:support-set-expansion}, we have $\mathsf{supp}(q_t^{\prime})\supseteq\mathsf{supp}(q_t)$ for any $t\in(0,1)$, meaning that $\mathbf{x}_t\in \mathsf{supp}(q_t^\prime)$.  
Hence we have $\{\mathbf{x}_t\}_{t\in[0,1]} \in \mathbb{T}^\prime$.
This directly implies $\mathbb{T}^{\prime}\supseteq\mathbb{T}$.
\end{proof}

\subsection{Proof of \cref{theorem:divergent-lower-bound-time-score}}

\begin{proof}\label{proof:divergent-lower-bound-time-score}
    For the DI, the time derivative of its log-density is governed by the Fokker-Planck equation:
\begin{equation}
\partial_t \log q_t = -\nabla \cdot \mathbf{u}_t - \mathbf{u}_t \cdot \nabla \log q_t,
\end{equation}
where $\nabla\cdot$ and $\nabla$ are the divergence and gradient operators w.r.t. $\mathbf{x}_t$. 
$\mathbf{u}_t(\mathbf{x}_t) = \mathbb{E}_{\pi}[\dot{\alpha}_t\mathbf{X}_0 + \dot{\beta}_t\mathbf{X}_1 \mid\mathbf{X}_t=\mathbf{x}_t ]$ is the drift term.
Taking expectations over $q_t$ and applying the triangle inequality and Cauchy-Schwarz inequality to this equation:
\begin{equation}
\begin{aligned}
    \mathbb{E}_{q_t}\left[\left|\partial_t \log q_t\right |\right] &=\mathbb{E}_{q_t}\left[|-\nabla \cdot \mathbf{u}_t - \mathbf{u}_t \cdot \nabla \log q_t|\right]  \\
    &\geq \mathbb{E}_{q_t}\left[\left||\nabla \cdot \mathbf{u}_t| - |\mathbf{u}_t \cdot \nabla \log q_t|\right|\right]  \quad \left( \text{triangle inequality}\right) \\
    &\geq \mathbb{E}_{q_t}\left[|\nabla \cdot \mathbf{u}_t|\right] - \mathbb{E}_{q_t}\left[|\mathbf{u}_t \cdot \nabla \log q_t|\right]  \\
    &\geq \mathbb{E}_{q_t}[\left| \nabla \cdot \mathbf{u}_t \right| ] - \mathbb{E}_{q_t}[ \|\mathbf{u}_t\| \cdot \|\nabla \log q_t\|] \quad \left( \text{Cauchy-Schwarz inequality}\right)
\end{aligned}  \label{appendix:eq-lower-bound-time-score}
\end{equation}

\textbf{(1) Lower bound of $\mathbb{E}_{q_t}[\left| \nabla \cdot \mathbf{u}_t \right| ]$.} 
The divergence term $\nabla \cdot \mathbf{u}_t$ is computed via the Jacobian of the inverse mapping:
\begin{equation}
    \begin{aligned}
        \nabla \cdot \mathbf{u}_t &= \text{tr}(\nabla \mathbf{u}_t)= \text{tr}\left( \mathbb{E}_{\pi}\left[ \dot{\alpha}_t \nabla_{\mathbf{x}_t}\mathbf{X}_0 + \dot{\beta}_t \nabla_{\mathbf{x}_t}\mathbf{X}_1 \mid\mathbf{X}_t=\mathbf{x}_t \right] \right)\\
        &=\mathbb{E}_{\pi}\left[\dot{\alpha}_t\text{tr}\left( \nabla_{\mathbf{x}_t}\mathbf{X}_0 \right) + \dot{\beta}_t \text{tr}\left( \nabla_{\mathbf{x}_t}\mathbf{X}_1  \right) \mid \mathbf{x}_t \right]. 
    \end{aligned}
\end{equation}

For a given sample $(\mathbf{x}_0,\mathbf{x}_1)\sim\pi$, differentiating the the interpolation constraint $\mathbf{x}_t = \alpha_t\mathbf{x}_0 + \beta_t\mathbf{x}_1$ implicitly gives $\mathbf{E}_d = \alpha_t \nabla_{\mathbf{x}_t}\mathbf{x}_0 + \beta_t \nabla_{\mathbf{x}_t}\mathbf{x}_1$. 
Rearranging yields $
\nabla_{\mathbf{x}_t}\mathbf{x}_0 = \alpha_t^{-1}\mathbf{E}_d - \beta_t\alpha_t^{-1}\nabla_{\mathbf{x}_t}\mathbf{x}_1$ and $ \nabla_{\mathbf{x}_t}\mathbf{x}_1 = \beta_t^{-1}\mathbf{E}_d - \alpha_t\beta_t^{-1}\nabla_{\mathbf{x}_t}\mathbf{x}_0$.

Decomposing $\nabla_{\mathbf{x}_t}\mathbf{x}_0$ into its independent case value and a residual:
\begin{equation}
\nabla_{\mathbf{x}_t}\mathbf{x}_0 = \alpha_t^{-1}\mathbf{E}_d + \mathbf{R}_t,
\end{equation}
where $\mathbf{R}_t = -\beta_t\alpha_t^{-1}\nabla_{\mathbf{x}_t}\mathbf{x}_1$ denote the residual term in the Jacobian decomposition. 
By \cref{assumption:Lip-conditional-distri} and the definition of the score function of the conditional distribution, $\nabla_{\mathbf{x}_t} \log q_0(\mathbf{x}_0\mid\mathbf{x}_t) = -\nabla_{\mathbf{x}_t}\mathbf{x}_0 \cdot \nabla_{\mathbf{x}_0} \log q_0(\mathbf{x}_0\mid\mathbf{x}_t)$, the Lipschitz continuity of the conditional scores implies $\|\nabla_{\mathbf{x}_t}\mathbf{x}_0\|_{\text{op}} \leq L\alpha_t^{-1}$ and $\|\nabla_{\mathbf{x}_t}\mathbf{x}_1\|_{\text{op}} \leq L\beta_t^{-1}$. Here $\|\cdot\|_{\text{op}}$ denotes the operator norm. 
Hence, the absolute value of the trace of the  residual term satisfies:
\begin{equation}
|\text{tr}(\mathbf{R}_t)| \leq d \cdot \|\mathbf{R}_t\|_{\text{op}} = d\alpha_t^{-1}\beta_t\|\nabla_{\mathbf{x}_t}\mathbf{X}_1\|_{\text{op}} \leq d\alpha_t^{-1}\beta_t \cdot \frac{L}{\beta_t} = dL\alpha_t^{-1}.
\end{equation}
This directly leads to the conditional expectation: $\left| \mathbb{E}_{q_t}[\text{tr}\left(\mathbf{R}_t\right)\mid\mathbf{X}_t] \right| \leq \mathbb{E}_{q_t}\left[ \left| \text{tr}\left(\mathbf{R}_t\right)\right|\mid\mathbf{X}_t\right] \leq dL\alpha_t^{-1}$.
Thus, the conditional expectation of $\text{tr}(\nabla_{\mathbf{x}_t}\mathbf{X}_0)$ satisfies:
\begin{equation}
\begin{aligned}
    \left|\mathbb{E}_{q_t}\left[\text{tr}(\nabla_{\mathbf{x}_t}\mathbf{X}_0)\mid\mathbf{X}_t\right] \right| &= \left|\mathbb{E}_{q_t}\left[\text{tr}\left(\alpha_t^{-1}\mathbf{E}_d + \mathbf{R}_t\right) \mid\mathbf{X}_t \right] \right|= \left|\mathbb{E}_{q_t}\left[d\alpha_t^{-1} +\text{tr}\left( \mathbf{R}_t\right)\mid\mathbf{X}_t\right] \right| \\
    &=\left| d\alpha_t^{-1} + \mathbb{E}_{q_t}[\text{tr}\left(\mathbf{R}_t\right)\mid\mathbf{X}_t] \right|  \\
    &\geq d\alpha_t^{-1} - \left| \mathbb{E}_{q_t}[\text{tr}\left(\mathbf{R}_t\right)\mid\mathbf{X}_t] \right| \\
    &\geq d\alpha_t^{-1} - dL\alpha_t^{-1} = d\alpha_t^{-1}(1-L).
    % &= \left| \frac{d}{\alpha_t} + \text{tr}(\mathbb{E}_{\Pi}[\mathbf{R}_t]) \right| \\
    % &\geq \frac{d}{\alpha_t} - \frac{dL}{\alpha_t}\mathcal{W}_2(q_0,q_1) \\
    % &=\frac{d}{\alpha_t}(1-L\mathcal{W}_2(q_0,q_1)).
\end{aligned}
\end{equation}

Similarly, we have $\left|\mathbb{E}_{q_t}[\text{tr}(\nabla_{\mathbf{x}_t}\mathbf{X}_1)\mid\mathbf{x}_t] \right| \geq d\beta_t^{-1}(1-L)$.
Base on these lower bounds and applying the triangle inequality, we have:
\begin{equation}
\begin{aligned}
    \mathbb{E}_{ q_t}\left[\left| \nabla \cdot \mathbf{u}_t \right| \right]&=\mathbb{E}_{ q_t}\left[\left | \mathbb{E}_{q_t}\left[\dot{\alpha}_t\text{tr}\left( \nabla_{\mathbf{x}_t}\mathbf{X}_0 \right) + \dot{\beta}_t \text{tr}\left( \nabla_{\mathbf{x}_t}\mathbf{X}_1 \right) \mid\mathbf{X}_t\right] \right | \right] \\
    &\geq \mathbb{E}_{q_t}\left[|\dot{\alpha}_t| \left | \mathbb{E}_{q_t}\left[ \text{tr}\left( \nabla_{\mathbf{x}_t}\mathbf{X}_0 \right) \mid\mathbf{X}_t\right] \right| - |\dot{\beta}_t| \left | \mathbb{E}_{q_t}\left[ \text{tr}\left( \nabla_{\mathbf{x}_t}\mathbf{X}_1 \right) \mid\mathbf{X}_t\right] \right| \right] \\
    &\geq \mathbb{E}_{q_t}\left[d|\dot{\alpha}_t|\alpha_t^{-1}(1-L) - |\dot{\beta}_t| \mathbb{E}_{q_t}\left[ \left| \text{tr}\left( \nabla_{\mathbf{x}_t}\mathbf{X}_1 \right) \right|\mid\mathbf{X}_t \right] \right] \\
    & \geq \mathbb{E}_{q_t}\left[d|\dot{\alpha}_t|\alpha_t^{-1}(1-L) - |\dot{\beta}_t|\cdot dL\beta_t^{-1} \right] \quad \left( \left| \text{tr}\left( \nabla_{\mathbf{x}_t}\mathbf{X}_1 \right) \right| \leq d\cdot \|\nabla_{\mathbf{x}_t}\mathbf{X}_1\|_{\text{op}}\leq dL\beta_t^{-1} \right)  \\
    &=d\left( (1-L) |\dot{\alpha}_t|\alpha_t^{-1} - L|\dot{\beta}_t|\beta_t^{-1} \right).
\end{aligned} \label{appendix:eq-lower-bound-div-ut}
\end{equation}

\textbf{(2) Upper bound of $\mathbb{E}_{q_t}[\|\mathbf{u}_t\| \cdot \|\nabla \log q_t\|]$.} 
Applying Cauchy-Schwarz:
\begin{equation}
\begin{aligned}
    \mathbb{E}_{q_t}[\|\mathbf{u}_t\| \cdot \|\nabla \log q_t\|] &\leq \sqrt{ \mathbb{E}_{q_t}[\|\mathbf{u}_t\|^2] } \cdot \sqrt{ \mathbb{E}_{q_t}[\|\nabla \log q_t\|^2] } \\
    &= \sqrt{ \mathbb{E}_{q_t}\left[\left\|\mathbb{E}_{\pi}\left[ \dot{\alpha}_t\mathbf{X}_0 + \dot{\beta}_t\mathbf{X}_1 \big|\mathbf{X}_t\right]\right\| ^2\right] } \cdot \sqrt{ \mathbb{E}_{q_t}[\|\nabla \log q_t\|^2] }  \\
    &\leq \sqrt{ \mathbb{E}_{q_t}\left[\mathbb{E}_{\pi}\left[ \|\dot{\alpha}_t\mathbf{X}_0 + \dot{\beta}_t\mathbf{X}_1\| ^2 \big|\mathbf{X}_t\right]\right] } \cdot \sqrt{ \mathbb{E}_{q_t}[\|\nabla \log q_t\|^2] } \\
    &\leq \sqrt{ 2\dot{\alpha}_t^2 \mathbb{E}_{q_0}[\|\mathbf{X}_0\|^2] + 2\dot{\beta}_t^2 \mathbb{E}_{q_1}[\|\mathbf{X}_1\|^2] } \cdot \sqrt{ \mathbb{E}_{q_t}[\|\nabla \log q_t\|^2] } =\mathcal{O}\left(\sqrt{ \mathbb{E}_{q_t}[\|\nabla \log q_t\|^2] }\right).
\end{aligned}  \label{appendix:eq-upper-bound-residual}
\end{equation}
Here the drift norm $\mathbb{E}_{q_t}[\|\mathbf{u}_t\|^2]$ is bounded by the second moments of $\mathbf{X}_0$ and $\mathbf{X}_1$, which are finite under our assumptions.

\textbf{(3) Lower bound of $\mathbb{E}_{q_t}\left[\left|\partial_t \log q_t\right |\right]$.} Finally, bring \cref{appendix:eq-lower-bound-div-ut} and \cref{appendix:eq-upper-bound-residual} back to \cref{appendix:eq-lower-bound-time-score}, the lower bound of $\mathbb{E}_{q_t}\left[\left|\partial_t \log q_t\right |\right]$ can be derived:
\begin{equation}
\begin{aligned}
    \mathbb{E}_{q_t}\left[\left|\partial_t \log q_t\right |\right] &\geq \mathbb{E}_{q_t}[\left| \nabla \cdot \mathbf{u}_t \right| ] - \mathbb{E}_{q_t}[ \|\mathbf{u}_t\| \cdot \|\nabla \log q_t\|] \\
    &\geq \underbrace{d\left( (1-L) \frac{|\dot{\alpha}_t|}{\alpha_t} - L\frac{|\dot{\beta}_t|}{\beta_t} \right)}_{\text{Divergence term}} - \underbrace{\mathcal{O}\left(\sqrt{ \mathbb{E}_{q_t}[\|\nabla \log q_t\|^2] }\right)}_{\text{Residual term}}.
\end{aligned}
\end{equation}

The residual term $\mathcal{O}(\sqrt{\mathbb{E}_{q_t}[\|\nabla \log q_t\|^2]})$ is finite under \cref{assumption:Lip-conditional-distri} (Lipschitz scores imply bounded Fisher information). Hence, the divergence term dominates asymptotically. %\textcolor{red}{$\|\nabla \log q_t\|^2$ can be evaluated under Gaussian assumptions, $$\log q_t(\mathbf{x}) = -\frac{d}{2} \log(2\pi) - \frac{d}{2} \log \sigma_t^2 - \frac{1}{2\sigma_t^2} \|\mathbf{x} - \boldsymbol{\mu}_t\|^2.$$}

Near the boundaries $t \to 0^+$ and $t \to 1^-$, the terms $\frac{|\dot{\alpha}_t|}{\alpha_t}$ and $\frac{|\dot{\beta}_t|}{\beta_t}$ dominate due to the monotonicity and boundary conditions: 
\begin{equation}
    \begin{aligned}
        \text{(1) As }& t \to 0^+: 
  \alpha_t \to 1, \quad \beta_t \to 0, \quad \frac{|\dot{\alpha}_t|}{\alpha_t} \sim |\dot{\alpha}_0|, \quad \frac{|\dot{\beta}_t|}{\beta_t} \sim \frac{\dot{\beta}_0^+}{\beta_t} \to +\infty. \\
  \text{(2) As }& t \to 1^-: 
  \alpha_t \to 0, \quad \beta_t \to 1, \quad \frac{|\dot{\alpha}_t|}{\alpha_t} \sim \frac{|\dot{\alpha}_1^-|}{\alpha_t} \to +\infty, \quad \frac{|\dot{\beta}_t|}{\beta_t} \sim |\dot{\beta}_1|.
    \end{aligned}
\end{equation}

For any $L<1$,  the prefactor $1 - L > 0$ ensures:
\begin{equation}
    \begin{aligned}
        \lim_{t \to 1^-} \mathbb{E}_{q_t}\left[\|\partial_t \log q_t\| \right] \geq \lim_{t \to 1^-} d(1 - L)\frac{|\dot{\alpha}_t|}{\alpha_t} = +\infty.
    \end{aligned}
\end{equation}
This concludes the universal boundary divergence.

\end{proof}

\subsection{Proof of \cref{proposition:uniform-estimation-formula}}
\label{proof:uniform-estimation-formula}
\begin{proof}
    Assume both $ q_0$ and $ q_1$ are smooth with sufficient differentiability. The Gaussian-dequantified densities are given by:
    \begin{equation}
    \begin{aligned}
        q_i^\prime(\mathbf{x}) &= (q_i \ast \mathcal{N}(\mathbf{0}, \varepsilon \mathbf{E}_d))(\mathbf{x}) = \int q_i(\mathbf{x}^\prime) \mathcal{N}(\mathbf{x} ; \mathbf{x}^\prime, \varepsilon \mathbf{E}_d) \mathrm{d}\mathbf{x}^\prime  \\
        &=\int q_i(\mathbf{x}^\prime) \left[ \delta(\mathbf{x} - \mathbf{x}^\prime) + \frac{\varepsilon}{2} \nabla_{\mathbf{x}^\prime}^2 \delta(\mathbf{x} - \mathbf{x}^\prime)+ \mathcal{O}(\varepsilon^2) \right] \mathrm{d}\mathbf{x}^\prime \quad (\text{Taylor expansion around }\mathbf{x}^\prime) \\
        &=q_i(\mathbf{x})+\frac{\varepsilon}{2}\int q_i(\mathbf{x}^{\prime}) \nabla_{\mathbf{x}^{\prime}}^2 \delta(\mathbf{x} - \mathbf{x}^{\prime})  \mathrm{d}\mathbf{x}^{\prime}+ \mathcal{O}(\varepsilon^2) \\
        &=q_i(\mathbf{x}) + \frac{\varepsilon}{2} \nabla_{\mathbf{x}}^2 q_i(\mathbf{x}) + \mathcal{O}(\varepsilon^2), \quad (\text{Integration by parts}) ,
    \end{aligned}\label{eq:convolution-integral}
    \end{equation}
    where $ \nabla_{\mathbf{x}^\prime}^2 $ is the Laplacian operator, $ \delta(\mathbf{x} - \mathbf{x}^\prime) $ is the Dirac delta. 

    Substituting these two expansions into the dequantified density ratio $ r^\prime(\mathbf{x}) = \frac{q_1^\prime(\mathbf{x})}{q_0^\prime(\mathbf{x})} $, we have:
    \begin{equation}
    \begin{aligned}
        r^\prime(\mathbf{x}) &= \frac{q_1^\prime(\mathbf{x})}{q_0^\prime(\mathbf{x})} = \frac{q_1(\mathbf{x}) + \frac{\varepsilon}{2} \nabla_{\mathbf{x}}^2 q_1(\mathbf{x})+\mathcal{O}(\varepsilon^2)}{q_0(\mathbf{x}) + \frac{\varepsilon}{2} \nabla_{\mathbf{x}}^2 q_0(\mathbf{x})+\mathcal{O}(\varepsilon^2)} \\
        &=\frac{q_1(\mathbf{x})}{q_0(\mathbf{x})} + \frac{\varepsilon}{2} \frac{\nabla_{\mathbf{x}}^2 q_1(\mathbf{x})}{q_0(\mathbf{x})} - \frac{\varepsilon}{2} r(\mathbf{x}) \frac{\nabla_{\mathbf{x}}^2 q_0(\mathbf{x})}{q_0(\mathbf{x})} + \mathcal{O}(\varepsilon^2) \quad (\text{First-order expansion of a fraction}) \\
        &=r(\mathbf{x})+\frac{\varepsilon}{2} \underbrace{\left[ \frac{\nabla_{\mathbf{x}}^2 q_1(\mathbf{x})}{q_0(\mathbf{x})} - r(\mathbf{x}) \frac{\nabla_{\mathbf{x}}^2 q_0(\mathbf{x})}{q_0(\mathbf{x})} \right]}_{\Delta(\mathbf{x})} + \mathcal{O}(\varepsilon^2).
    \end{aligned}
    \end{equation}

    To bound $ r^\prime(\mathbf{x}) - r(\mathbf{x}) $ in $L^\infty$, the supremum can be computed under \cref{ass:smoothness}:
    \begin{equation}
    \begin{aligned}
        \| r^\prime - r \|_{L^\infty} &\leq  \frac{\varepsilon}{2} \sup_{\mathbf{x}} |\Delta(\mathbf{x})| + \mathcal{O}(\varepsilon^2) \\
        &= \frac{\varepsilon}{2} \sup_{\mathbf{x}} \left| \frac{\nabla_{\mathbf{x}}^2 q_1(\mathbf{x})}{q_0(\mathbf{x})} - r(\mathbf{x}) \frac{\nabla_{\mathbf{x}}^2 q_0(\mathbf{x})}{q_0(\mathbf{x})} \right| + \mathcal{O}(\varepsilon^2) \\
        &\leq \frac{\varepsilon}{2} \sup_{\mathbf{x}} \left| \frac{\nabla_{\mathbf{x}}^2 q_1(\mathbf{x})}{q_0(\mathbf{x})} \right| + \frac{\varepsilon}{2} \sup_{\mathbf{x}} \left| r(\mathbf{x}) \frac{\nabla_{\mathbf{x}}^2 q_0(\mathbf{x})}{q_0(\mathbf{x})} \right| + \mathcal{O}(\varepsilon^2) \\
        &\leq \frac{\varepsilon}{2} \left( \frac{\|\nabla_{\mathbf{x}}^2 q_1\|_{L^\infty}}{\inf\limits_{\mathbf{x}} q_0(\mathbf{x})} + \frac{ \|\nabla_{\mathbf{x}}^2 q_0\|_{L^\infty}}{\inf\limits_{\mathbf{x}} q_0(\mathbf{x})} \sup_{\mathbf{x}} r(\mathbf{x})  \right) + \mathcal{O}(\varepsilon^2) \\
        &=\frac{\varepsilon}{2\inf\limits_{\mathbf{x}} q_0(\mathbf{x})} \left( \|\nabla_{\mathbf{x}}^2 q_1\|_{L^\infty} + \|\nabla_{\mathbf{x}}^2 q_0\|_{L^\infty}\|r\|_{L^\infty} \right) + \mathcal{O}(\varepsilon^2). 
    \end{aligned}
    \end{equation}
    
    Under \cref{ass:smoothness}, $\inf\limits_{\mathbf{x}} q_0(\mathbf{x})$ has lower bound and these norms have upper bound. Then as $\varepsilon \to 0$,
    \begin{equation}
        \| r^\prime - r \|_{L^\infty} \leq \mathcal{O}(\varepsilon),
    \end{equation}
    where the constant  $ C $ is given by: $C = \frac{\|\nabla_{\mathbf{x}}^2 q_1\|_{L^\infty} + \|\nabla_{\mathbf{x}}^2 q_0\|_{L^\infty} \|r\|_{L^\infty}}{2\inf\limits_{\mathbf{x}} q_0(\mathbf{x})}.$
    Hence, as $\varepsilon\to 0$, we have $ r^\prime(\mathbf{x}) \to r(\mathbf{x}) $, verifying the stated proposition.
\end{proof}

\subsection{Proof of \cref{proposition:solution-to-SB-problem}}
\label{proof:solution-to-SB-problem}

\begin{proof}
    Based on Theorem 2.4 of \citep{leonard2014some}, \citep{de2021diffusion} established that the solution to the SB problem as detailed in \cref{eq:SB-problem}, $\pi^{\star}$, satisfies the SB conditions: 
    (1) the optimization problem for $\pi^{\star}$ is equivalent to the entropically regularied OT problem, with the optimal coupling $\pi _{2\gamma^2}$ defined in \cref{eq:entropic-OT-problem}; 
    (2) for samples $\left(\mathbf{x}_0^\prime,\mathbf{x}_1^\prime\right)\sim\pi^\star$, the associated  conditional path distributions $\pi^{\star}(\cdot\mid \mathbf{x}_0^\prime,\mathbf{x}_1^\prime)$ minimize  the KL divergence: $\mathbb{E}_{(\mathbf{x}_0^\prime,\mathbf{x}_1^\prime)\sim\pi^\star}\mathrm{KL}\left(\pi^\star(\cdot\mid \mathbf{x}_0^\prime,\mathbf{x}_1^\prime)\|\pi_\mathrm{ref}(\cdot\mid \mathbf{x}_0^\prime,\mathbf{x}_1^\prime)\right)$, where $\pi_\mathrm{ref}$ is the reference path distribution satisfying $\log \pi_{\mathrm{ref}}( \mathbf{x}_0^\prime, \mathbf{x}_1^\prime) = \frac{\|\mathbf{x}_0^\prime-\mathbf{x}_1^\prime\|^2}{2\gamma^2}+ \text{const}$.
    These conditional distributions are optimized using Brownian bridges of diffusion scale $\gamma$, conditioned on the endpoints $\mathbf{x}_0^\prime$ and $\mathbf{x}_1^\prime$.
    The marginal distribution at intermediate time $t$ along the Brownian bridge is given by $q_t(\mathbf{x}\mid \mathbf{x}_0^\prime, \mathbf{x}_1^\prime)=\mathcal{N}(\mathbf{x}\mid (1-t)\mathbf{x}_0^\prime + t\mathbf{x}_1^\prime,t(1-t)\gamma^2\mathbf{E}_d)$ \citep{tong2023improving}.
    Since our proposed OTR method uses   Sinkhorn's algorithm to solve the entropically regularized OT problem, and the probability paths of our DDBI with $\alpha_t=1-t$ and $\beta_t=t$ align  with those of the Brownian bridge, a trajectory $\mathbf{x}_t^\prime$ generated by first  sampling $(\mathbf{x}_0^\prime, \mathbf{x}_1^\prime) \sim \pi^\star$, then sampling $\mathbf{x}_t \sim q_t(\cdot \mid \mathbf{x}_0^\prime, \mathbf{x}_1^\prime)$ satisfies the SB conditions, thus verifying the proposition.
\end{proof}

\subsection{Proof of \cref{theorem:upper-bound-of-DDBI-DSBI}}
\label{proof:upper-bound-of-DDBI-DSBI}

\begin{proof}

To analyze $\mathrm{Var}_{q_t^\prime}(\partial_t \log q_t^\prime)$, applying the law of total variance to $\mathrm{Var}_{q_t^\prime}(\partial_t \log q_t^\prime)$:
\begin{equation}  \label{appendix:eq-total-variance-law}
    \mathrm{Var}_{q_t^\prime}(\partial_t \log q_t^\prime) = \mathbb{E}_{( \mathbf{X}_0^\prime, \mathbf{X}_1^\prime ) \sim \pi}\left[\mathrm{Var}\left(\partial_t \log q_t^\prime(\mathbf{X}_t^\prime \mid \mathbf{X}_0^\prime, \mathbf{X}_1^\prime)\right)\right] + \mathrm{Var}_{( \mathbf{X}_0^\prime, \mathbf{X}_1^\prime ) \sim \pi}\left(\mathbb{E}\left[\partial_t \log q_t^\prime(\mathbf{X}_t^\prime \mid \mathbf{X}_0^\prime, \mathbf{X}_1^\prime)\right]\right).
\end{equation}

For any paired endpoints $(\mathbf{X}_0^\prime, \mathbf{X}_1^\prime) \sim \pi$, the interpolant $\mathbf{X}_t^\prime$ follows a Gaussian distribution conditioned on the endpoints $\mathbf{X}_t^\prime \mid (\mathbf{X}_0^\prime, \mathbf{X}_1^\prime) \sim q_t^\prime(\cdot\mid\mathbf{X}_0^\prime, \mathbf{X}_1^\prime)=\mathcal{N}\left( \boldsymbol{\mu}_t, \sigma_t^2 \mathbf{I}_d \right)$, where $\boldsymbol{\mu}_t = \alpha_t \mathbf{X}_0^\prime + \beta_t \mathbf{X}_1^\prime,  \sigma_t^2 = t(1-t)\gamma^2 + (\alpha_t^2 + \beta_t^2)\varepsilon$.
The conditional score $\partial_t \log q_t^\prime(\mathbf{X}_t^\prime\mid\mathbf{X}_0^\prime, \mathbf{X}_1^\prime)$ is derived explicitly as:
\begin{equation}\label{appendix:eq-conditional-time-score}
    \begin{aligned}
\partial_t \log q_t^\prime(\mathbf{X}_t^\prime\mid\mathbf{X}_0^\prime, \mathbf{X}_1^\prime) &=\frac{\partial}{\partial t} \log \mathcal{N}\left(\boldsymbol{\mu}_t, \sigma_t^2 \mathbf{I}_d \right)= \frac{\partial}{\partial t} \log \left[ (2\pi \sigma_t^2)^{-d/2} \exp\left(-\frac{\|\mathbf{X}_t^\prime - \boldsymbol{\mu}_t\|^2}{2\sigma_t^2}\right) \right] \\
&= - \frac{d}{2} \frac{\dot{\sigma}_t^2}{\sigma_t^2} +\frac{\dot{\alpha}_t \mathbf{X}_0^\prime + \dot{\beta}_t \mathbf{X}_1^\prime}{\sigma_t^2} \cdot (\mathbf{X}_t^\prime - \boldsymbol{\mu}_t) + \frac{\|\mathbf{X}_t^\prime - \boldsymbol{\mu}_t\|^2}{2} \frac{\dot{\sigma}_t^2}{\sigma_t^4} \\
&= -\frac{d \dot{\sigma}_t^2}{2\sigma_t^2} + \frac{\dot{\alpha}_t \mathbf{X}_0^\prime + \dot{\beta}_t \mathbf{X}_1^\prime}{\sigma_t} \cdot \mathbf{Z}_t + \frac{\dot{\sigma}_t^2 \|\mathbf{Z}_t\|^2}{2\sigma_t^2}. 
\end{aligned}
\end{equation}

\textbf{The term $\mathbb{E}_{\pi}\left[\mathrm{Var}\left(\partial_t \log q_t^\prime(\mathbf{X}_t^\prime \mid \mathbf{X}_0^\prime, \mathbf{X}_1^\prime)\right)\right]$.} 
Taking the variance on both sides of \cref{appendix:eq-conditional-time-score} and taking expectation over the coupling distribution $\pi$:
\begin{equation}\label{appendix:eq-exp-var-time-score}
    \begin{aligned}
\mathbb{E}_{\pi}\left[\mathrm{Var}(\partial_t \log q_t^\prime(\mathbf{X}_t^\prime \mid \mathbf{X}_0^\prime, \mathbf{X}_1^\prime))\right] &=\mathbb{E}_{\pi}\left[ \mathrm{Var}\left( \frac{\dot{\alpha}_t \mathbf{X}_0^\prime + \dot{\beta}_t \mathbf{X}_1^\prime}{\sigma_t} \cdot \mathbf{Z}_t \right) + \mathrm{Var}\left( -\frac{d \dot{\sigma}_t^2}{2\sigma_t^2} \right) + \mathrm{Var}\left( -\frac{\dot{\sigma}_t^2 \|\mathbf{Z}_t\|^2}{2\sigma_t^2} \right)\right] \\
&= \frac{1}{\sigma_t^2}\mathbb{E}_{\pi}\left[\left\|\dot{\alpha}_t \mathbf{X}_0^\prime + \dot{\beta}_t \mathbf{X}_1^\prime\right\|^2 + 0 + \frac{\dot{\sigma}_t^4}{4\sigma_t^4} \mathrm{Var}(\|\mathbf{Z}_t\|^2)\right] \\
&= \frac{1}{\sigma_t^2}\mathbb{E}_{\pi}\left[\left\|\dot{\alpha}_t \mathbf{X}_0^\prime + \dot{\beta}_t \mathbf{X}_1^\prime\right\|^2 \right] + \frac{\dot{\sigma}_t^4 d}{2\sigma_t^4}.
\end{aligned}
\end{equation}

Specifically, when $\alpha_t = 1-t$ and $\beta_t = t$, this reduces to $\mathbb{E}_{\pi}\left[\mathrm{Var}(\partial_t \log q_t^\prime(\mathbf{X}_t^\prime \mid \mathbf{X}_0^\prime, \mathbf{X}_1^\prime))\right]=\frac{1}{\sigma_t^2}\mathbb{E}_{\pi}\left[\left\| \mathbf{X}_0^\prime - \mathbf{X}_1^\prime\right\|^2 \right] + \frac{\dot{\sigma}_t^4 d}{2\sigma_t^4}$.

\textbf{The term $\mathrm{Var}\left(\mathbb{E}\left[\partial_t \log q_t^\prime(\mathbf{X}_t^\prime \mid \mathbf{X}_0^\prime, \mathbf{X}_1^\prime)\right]\right)$.}
Taking the variance on both sides of \cref{appendix:eq-conditional-time-score} and taking expectation over the coupling distribution $\pi$:
\begin{equation}\label{appendix:eq-var-exp-time-score}
    \begin{aligned}
        \mathrm{Var}_{\pi}\left(\mathbb{E}\left[\partial_t \log q_t^\prime(\mathbf{X}_t^\prime \mid \mathbf{X}_0^\prime, \mathbf{X}_1^\prime)\right]\right) &= \mathrm{Var}_{\pi}\left(\mathbb{E}\left[-\frac{d \dot{\sigma}_t^2}{2\sigma_t^2} + \frac{\dot{\alpha}_t \mathbf{X}_0^\prime + \dot{\beta}_t \mathbf{X}_1^\prime}{\sigma_t} \cdot \mathbf{Z}_t + \frac{\dot{\sigma}_t^2 \|\mathbf{Z}_t\|^2}{2\sigma_t^2}\right]\right) \\
        &=\mathrm{Var}_{\pi}\left(-\frac{d \dot{\sigma}_t^2}{2\sigma_t^2} + \frac{\dot{\alpha}_t \mathbf{X}_0^\prime + \dot{\beta}_t \mathbf{X}_1^\prime}{\sigma_t} \cdot \mathbb{E}[\mathbf{Z}_t] + \frac{\dot{\sigma}_t^2 }{2\sigma_t^2}\mathbb{E}\left[ \|\mathbf{Z}_t\|^2\right]\right)  \\
        &=\mathrm{Var}_{\pi}\left( \frac{\dot{\alpha}_t \mathbf{X}_0^\prime + \dot{\beta}_t \mathbf{X}_1^\prime}{\sigma_t} \cdot \mathbf{0} + \frac{\dot{\sigma}_t^2 }{2\sigma_t^2}d\right)=0.
    \end{aligned}
\end{equation}
The last equality holds because $\mathbf{Z}_t$ is a Gaussian noise, leading to $\mathbf{Z}_t\sim\mathcal{N}(\mathbf{0},\mathbf{E}_d)$ and $\|\mathbf{Z}_t\|^2\sim \chi(d)$.

Bringing \cref{appendix:eq-exp-var-time-score,appendix:eq-var-exp-time-score} into \cref{appendix:eq-total-variance-law}, the variance term $\mathrm{Var}_{q_t^\prime}(\partial_t \log q_t^\prime)$ becomes:
\begin{equation}
\begin{aligned}
    \mathrm{Var}_{q_t^\prime}(\partial_t \log q_t^\prime)=\int_0^1 \left[ \frac{1}{\sigma_t^2}\mathbb{E}_{\pi}\left[\left\|\mathbf{X}_0^\prime - \mathbf{X}_1^\prime\right\|^2 \right] + \frac{\dot{\sigma}_t^4 d}{2\sigma_t^4} \right] \mathrm{d}t =\mathbb{E}_{\pi}\left[\left\|\mathbf{X}_0^\prime - \mathbf{X}_1^\prime\right\|^2 \right] \int_0^1 \frac{1}{\sigma_t^2} \mathrm{d}t + \int_0^1 \frac{\dot{\sigma}_t^4 d}{2\sigma_t^4} \mathrm{d}t,
\end{aligned}
\end{equation}
when $\alpha_t = 1-t$ and $\beta_t = t$. 
Thus, the difference between the upper bound of the variance for DSBI and DDBI becomes:
\begin{equation}
\begin{aligned}
    &\mathrm{Var}^{\text{DDBI}}_{q_t^\prime}(\partial_t \log q_t^\prime) - \mathrm{Var}^{\text{DSBI}}_{q_t^\prime}(\partial_t \log q_t^\prime)\\
    =&\mathbb{E}_{\pi}\left[\left\|\mathbf{X}_0^\prime - \mathbf{X}_1^\prime\right\|^2 \right] - \mathbb{E}_{\pi_{2\gamma^2}}\left[\left\|\hat{\mathbf{X}}_0^\prime - \hat{\mathbf{X}}_1^\prime\right\|^2 \right]\\
    =&\left[ \mathbb{E}_{\pi}\left[\left\|\mathbf{X}_0^\prime - \mathbf{X}_1^\prime\right\|^2 \right] - 2\gamma^2 \mathcal{H}(\pi_{2\gamma^2}) \right] - \left[ \mathbb{E}_{\pi_{2\gamma^2}}\left[\left\|\hat{\mathbf{X}}_0^\prime - \hat{\mathbf{X}}_1^\prime\right\|^2 \right] - 2\gamma^2 \mathcal{H}(\pi_{2\gamma^2})\right]\geq 0.
\end{aligned}
\end{equation}
This inequality holds because $\pi_{2\gamma^2}$ is the solution to the entropically regularized OT problem. This completes the proof.

\end{proof}

\subsection{Proof of \cref{corollary:upper-bound-time-score-DDBI-DSBI}}
\label{proof:upper-bound-time-score-DDBI-DSBI}
\begin{proof}
Directly applying the Jensen's inequality to $\mathbb{E}_{q_t^\prime}[|\partial_t \log q_t^\prime|]$ yields:
\begin{equation}
    \begin{aligned}
    \mathbb{E}_{q_t^\prime}[|\partial_t \log q_t^\prime|] \leq \sqrt{\mathbb{E}_{q_t^\prime}[(\partial_t \log q_t^\prime)^2]} =\sqrt{\mathrm{Var}_{q_t^\prime}(\partial_t \log q_t^\prime) + \left( \mathbb{E}_{q_t^\prime}[\partial_t \log q_t^\prime] \right)^2}.  \quad (\text{Property of variance}).
\end{aligned}
\end{equation}

The first term on the r.h.s. is bounded according to \cref{theorem:upper-bound-of-DDBI-DSBI}, satisfying $\mathrm{Var}_{q_t^\prime}(\partial_t \log q_t^\prime)=\frac{1}{\sigma_t^2}\mathbb{E}_{\pi}\left[\left\|\mathbf{X}_0^\prime - \mathbf{X}_1^\prime\right\|^2 \right] + \frac{\dot{\sigma}_t^4 d}{2\sigma_t^4}$ with $\sigma_t^2 = t(1-t)\gamma^2 + (\alpha_t^2 + \beta_t^2)\varepsilon$.
The second term on the r.h.s. vanishes identically $\mathbb{E}_{q_t^\prime}[\partial_t \log q_t^\prime] = \int \frac{\partial_t q_t^\prime}{q_t^\prime} q_t^\prime  \mathrm{d}\mathbf{x} = \int \partial_t q_t^\prime  \mathrm{d}\mathbf{x} = \partial_t \left( \int q_t^\prime  \mathrm{d}\mathbf{x} \right) = \partial_t (1) = 0$.
Thus, the term $\mathbb{E}_{q_t^\prime}[|\partial_t \log q_t^\prime|]$ is bounded by:
\begin{equation}
    \mathbb{E}_{q_t^\prime}[|\partial_t \log q_t^\prime|] \leq \sqrt{\frac{1}{\sigma_t^2}\mathbb{E}_{\pi}\left[\left\|\mathbf{X}_0^\prime - \mathbf{X}_1^\prime\right\|^2 \right] + \frac{\dot{\sigma}_t^4 d}{2\sigma_t^4}}< \infty.
\end{equation}

\end{proof}

\subsection{Derivation of \cref{proposition:log-density-ratio}}
\label{proof:theorem-log-density-ratio}
\begin{proof}
Let $\{\mathbf{X}_t^\prime\}_{t\in[0,1]}$ be a DDBI. It has a transition kernel   $q_t^\prime(\mathbf{x}\mid\mathbf{x}_0,\mathbf{x}_1)$ and marginal probability density $q_t^\prime(\mathbf{x})$. 
By dividing the interval $[0,1]$ into $M$ discrete intervals, the log dequantified density ratio for a given point $\mathbf{x}$ can be derived:
\begin{equation}
    \begin{aligned}
        \log r^{\prime}(\mathbf{x})&=\log\frac{ q_1^{\prime}(\mathbf{x})}{ q_0^{\prime}(\mathbf{x})} =\log \frac{q^{\prime}_{1/M}(\mathbf{x})}{q_0^{\prime}(\mathbf{x})} \frac{q_{2/M}^{\prime}(\mathbf{x})}{q_{1/M}^{\prime}(\mathbf{x})} \cdots  \frac{q_{1}^{\prime}(\mathbf{x})}{q_{(M-1)/M}^{\prime}(\mathbf{x})}  =\sum_{m=0}^{M-1}\log \frac{q_{(m+1)/M}^{\prime}(\mathbf{x})}{q_{m/M}^{\prime}(\mathbf{x})}. \label{eq:appendix-logr-integrate}
    \end{aligned}
\end{equation}

According to the Taylor's formula, we have $\log(1+\mathbf{x})\approx\mathbf{x}$ while $\mathbf{x}$ approaches 0. In this case, while $M$ is large enough so that the difference between $p_{m/M}(\mathbf{x}\mid \mathbf{x}_0, \mathbf{x}_1)$ and $ p_{(m-1)/M}(\mathbf{x}\mid \mathbf{x}_0, \mathbf{x}_1)$ approaches 0, we have
\begin{equation}
    \begin{aligned}
        \log \frac{q_{(m+1)/M}^{\prime}(\mathbf{x})}{q_{m/M}^{\prime}(\mathbf{x})}=\log\left( 1 +  \frac{q_{(m+1)/M}^{\prime}(\mathbf{x}) - q_{m/M}^{\prime}(\mathbf{x})}{q_{m/M}^{\prime}(\mathbf{x})} \right) \approx \frac{q_{(m+1)/M}^{\prime}(\mathbf{x}) - q_{m/M}^{\prime}(\mathbf{x})}{q_{M}^{\prime}(\mathbf{x})}.  
    \end{aligned}
\end{equation}
	
In the limit as $M\to\infty$, the difference term  $\frac{q_{(m+1)/M}^{\prime}(\mathbf{x}) - q_{m/M}^{\prime}(\mathbf{x})}{q_{M}^{\prime}(\mathbf{x})}$ can be seen as the approximation of $\frac{\partial}{\partial\tau} \log q_{\tau}^\prime(\mathbf{x})$ evaluated at $\tau=m/M$. 
Taking the limit as $M\to\infty$ for both sides of Eq. (\ref{eq:appendix-logr-integrate}), we can derive
\begin{equation}
    \begin{aligned}
        \log r^\prime(\mathbf{x})&=\lim\limits_{M\to\infty} \sum_{m=0}^{M-1}\log \frac{q_{(m+1)/M}^{\prime}(\mathbf{x})}{q_{m/M}^{\prime}(\mathbf{x})}   \\
        &\approx \lim\limits_{M\to\infty} \sum_{m=0}^{M-1} \frac{q_{(m+1)/M}^{\prime}(\mathbf{x}) - q_{m/M}^{\prime}(\mathbf{x})}{q_{M}^{\prime}(\mathbf{x})}   \\
        &\approx\lim\limits_{M\to\infty} \sum_{m=0}^{M-1} \left.\frac{\partial}{\partial\tau} \log q_{\tau}^\prime(\mathbf{x})\right|_{\tau=m/M} \\
        &=\int^{1}_{0}\partial_t \log q_{\tau}^\prime(\mathbf{x}) \mathrm{d}t. \\
    \end{aligned}
\end{equation}

According to \cref{proposition:uniform-estimation-formula}, the density ratio $r^\prime(\mathbf{x})$ uniformly approximates the target density ratio $r^{\star}(\mathbf{x})$, i.e. 
\begin{equation}
    \log r (\mathbf{x}) \approx \log r^\prime(\mathbf{x})\approx \int^{1}_{0}\partial_t \log q_{\tau}^\prime(\mathbf{x}) \mathrm{d}t.
\end{equation}
This completes the proof of this proposition. 
\end{proof}

\newpage
\section{Preliminaries}
\subsection{Special Cases of DI}
\label{appendix:special-cases-DI}
In the case of TRE, the interpolation strategy, as detailed in Eq. (\ref{eq:interpolant-TRE}), represents a specific case of $\mathbf{I}$, characterized by  $\alpha_t=\sqrt{1-\eta_t^2}, \beta_t=\eta_t$, with $t$ taking discrete values $0, 1/M, 2/M, \ldots, 1$. 
For DRE-$\infty$, the coefficients are defined as $\alpha_t=\exp\{-0.25(\beta_{\max}-\beta_{\min})t^2-0.5\beta_{\min}t\}$ and $\beta_t=\sqrt{1-\alpha_t^2}$ for the MNIST dataset and $\alpha_t = 1 - t$ and $\beta_t = t$ for other datasets.
The corresponding stochastic process for the former one aligns with the solution to variance preserving (VP) SDEs \citep{song2020score,li2024deepar,xin2024v,zhou2025hademif}. 

\subsection{From Optimal Transport to Entropic Regularization}
\label{appendix:optimal-transport}
The static OT problem seeks to find a coupling $\pi$ between two probability distributions $q_0$ and $q_1$ that minimizes a given cost function. For the 2-Wasserstein distance with a Euclidean ground cost $c(\mathbf{x}_0, \mathbf{x}_1) = \|\mathbf{x}_0 - \mathbf{x}_1\|^2$, the optimization problem is given by:
\begin{equation}\label{eq:transport-loss}
\mathcal{W}_2^2(q_0, q_1) = \inf_{\pi \in \Pi(q_0, q_1)} \int_{\mathbb{R}^d \times \mathbb{R}^d} \|\mathbf{x}_0 - \mathbf{x}_1\|^2 \mathrm{d}\pi(\mathbf{x}_0, \mathbf{x}_1),
\end{equation}
where $\Pi(q_0, q_1)$ denotes the set of joint probability measures with marginals $q_0$ and $q_1$.
The optimal solution for compactly supported distributions \citep{villani2009optimal} is characterized by straight-line interpolations between samples:
\begin{equation}
\mathbf{X}_t = (1-t)\mathbf{X}_0 + t\mathbf{X}_1, \quad t \in [0, 1],
\end{equation}
where $\mathbf{X}_0 \sim q_0$ and $\mathbf{X}_1 \sim q_1$. This interpolation aligns with the Benamou-Brenier formulation \citep{mccann1997convexity,zhou2024enhancing}, where the transport paths minimize the kinetic energy in the space of probability measures.

The natural connections between optimal transport theory and straight-line interpolations motivate the concept of Batch Optimal Transport (BatchOT) \citep{pooladian2023multisample,zhao2025less}. BatchOT provides a pseudo-deterministic coupling mechanism by extending the OT principles to minibatch sampling. This ensures practical scalability and aligns theoretical transport paths with computational requirements.

Despite its theoretical elegance, solving the OT problem at scale is computationally challenging due to its cubic complexity in the number of samples. 
Entropic regularization alleviates this issue by introducing an entropy penalty:
\begin{equation}
\mathcal{W}_{2,\xi}^2(q_0, q_1) = \inf_{\pi \in \Pi(q_0, q_1)} \int_{\mathbb{R}^d \times \mathbb{R}^d} \|\mathbf{x}_0 - \mathbf{x}_1\|^2 \mathrm{d}\pi(\mathbf{x}_0, \mathbf{x}_1) - \xi \mathcal{H}(\pi),
\end{equation}
where $\xi > 0$ is the regularization parameter and $\mathcal{H}(\pi)$ denotes the entropy of $\pi$. 
This formulation ensures convexity and allows scalable computation via Sinkhorn's algorithm \citep{cuturi2013sinkhorn}.

Entropic regularization connects OT with the Schr{\"o}dinger bridge (SB) problem, which models stochastic interpolation between distributions. Given a reference Wiener process scaled by $\gamma$, the SB problem finds the most probable stochastic process $\pi$ that satisfies the marginal constraints $q_0$ and $q_1$:
\begin{equation}
\pi^{\star} = \underset{\pi\in \Pi(q_0,q_1)}{\text{argmin}} \mathrm{KL}(\pi \parallel \pi_{\mathrm{ref}}),
\end{equation}
where $\pi_{\mathrm{ref}}$ is a reference process.
The SB solution corresponds to an entropy-regularized OT plan with $\xi = 2\gamma^2$:
\begin{equation}
\mathbf{X}_t = (1-t)\mathbf{X}_0 + t\mathbf{X}_1 + \sqrt{t(1-t)\gamma^2}\mathbf{Z}_t,
\end{equation}
where $\mathbf{Z}_t \sim \mathcal{N}(\mathbf{0}, \mathbf{E}_d)$. This formulation introduces stochasticity into the transport paths, effectively modeling uncertainty and noise.

\section{Experimental Details and More Results}

\subsection{Comparison of the Trajectory Sets for Interpolation Strategies}
In this section, we provide a detailed comparison of interpolation strategies, specifically deterministic interpolant (DI), diffusion bridge interpolant (DBI), dequantified diffusion bridge interpolant (DDBI), and dequantified Schr{\"o}dinger bridge interpolant (DSBI). Their intermediate samples and corresponding distributions are visualized in \cref{fig:compare-trajectory}.

\textbf{(a) DI:} 
DI constrains intermediate samples to fixed linear paths between $q_0(\mathbf{x})$ and $q_1(\mathbf{x})$, resulting in narrow bands across the trajectory space (\cref{fig:DI-concate}).
While dense along the paths, DI severely limits support and fails to explore alternative trajectories, making it inflexible and unsuitable for diverse distributions.

\textbf{(b) DBI:}
DBI introduces stochasticity through Brownian Bridge noise, expanding support and enabling broader trajectory exploration (\cref{fig:DBI-concate}). 
Compared to DI, DBI provides greater coverage and variability while retaining tractability, reducing the rigidity of interpolation paths \citep{zhao2023learning}.

\textbf{(c) DDBI:}
Extending DBI, DDBI modulates the noise with deterministic interpolation weights and diffusion components. This results in more dispersed trajectories (\cref{fig:DDBI-concate}) and a larger coverage of the intermediate distributions, balancing controlled stochasticity with enhanced flexibility \citep{zhou2025valuing}.

\textbf{(d) DSBI:} 
DSBI offers full stochastic control over noise and leverages entropy-regularized optimal transport, resulting in widely dispersed trajectories and efficient utilization of the trajectory space (\cref{fig:DSBI-concate}). By minimizing transition loss, DSBI achieves the largest support set and highest diversity among the methods, producing rich intermediate distributions.

Overall, our proposed methods (DBI, DDBI and DSBI) demonstrate clear advantages over deterministic baselines by achieving more comprehensive trajectory space exploration and flexible intermediate distribution generation. These results are consistent with our theoretical findings on support set and path set expansion, as formalized in \cref{theorem:support-set-expansion} and \cref{corollary:path-set-expansion}.

\subsection{Joint Score Matching}
\label{appendix:joint-score-matching}
In this section, we integrate the time score $s_{\boldsymbol{\theta}}^{t}\in\mathbb{R}$ and data score $\mathbf{s}_{\boldsymbol{\theta}}^{\mathbf{x}}\in\mathbb{R}^d$ to formulate the joint score $\mathbf{s}_{\boldsymbol{\theta}}^{t, \mathbf{x}}: [s_{\boldsymbol{\theta}}^{t}, \mathbf{s}_{\boldsymbol{\theta}}^{\mathbf{x}}]\in\mathbb{R}^{d+1}.$
This joint score is incorporated into the training objective defined in \cref{eq:time-score-matching-loss-L4}, resulting in a joint score matching objective \citep{choi2022density}:
\begin{equation}
\begin{aligned}
\mathcal{L}_{4}(\boldsymbol{\theta})&=2\mathbb{E}_{\mathbf{x}\sim q_{0}^\prime(\mathbf{x})}[\lambda(0)\mathbf{s}_{\boldsymbol{\theta}}^{t, \mathbf{x}}(\mathbf{x},0)[t]]-2\mathbb{E}_{\mathbf{x}\sim q_{1}^\prime(\mathbf{x})}[\lambda(1)\mathbf{s}_{\boldsymbol{\theta}}^{t, \mathbf{x}}(\mathbf{x},1)[t]] \\
&+\mathbb{E}_{t\sim q(t)}\mathbb{E}_{\mathbf{x}\sim q_{t}^\prime(\mathbf{x})}\mathbb{E}_{\mathbf{v}\sim q(\mathbf{v})}\left[2\lambda(t)\partial_t\mathbf{s}_{\boldsymbol{\theta}}^{t, \mathbf{x}}(\mathbf{x},t)[t]
+2\lambda^{\prime}(t)\mathbf{s}_{\boldsymbol{\theta}}^{t, \mathbf{x}}(\mathbf{x},t)[t] \right.\\
&\quad \left.+\lambda(t)\|\mathbf{s}_{\boldsymbol{\theta}}^{t, \mathbf{x}}(\mathbf{x},t)[\mathbf{x}]\|_2^2
+2\lambda(t)\mathbf{v}^{\mathsf{T}}\nabla_{\mathbf{x}}\mathbf{s}_{\boldsymbol{\theta}}^{t, \mathbf{x}}(\mathbf{x},t)[\mathbf{x}]\mathbf{v}\right],
\end{aligned}   \label{eq:time-score-matching-loss-L5}
\end{equation}
where $\mathbf{v}\sim q(\mathbf{v})=\mathcal{N}(\mathbf{0},\mathbf{E}_d)$ follows a standard Gaussian distribution, the terms $\mathbf{s}_{\boldsymbol{\theta}}^{t, \mathbf{x}}(\mathbf{x},t)[\mathbf{x}]$ and  $\mathbf{s}_{\boldsymbol{\theta}}^{t, \mathbf{x}}(\mathbf{x},t)[t]$ represent the data and time score components of $\mathbf{s}_{\boldsymbol{\theta}}^{t, \mathbf{x}}(\mathbf{x},t)$, respectively.

\subsection{Mutual Information Estimation}
\label{appendix:mutual-information}
Mutual information (MI) measures the dependency  between two random variables $\mathbf{X}\sim p(\mathbf{x})$ and $\mathbf{Y}\sim q(\mathbf{y})$, quantifying how much information one variable contains about the other. 
In this experiment, we employ $\text{D}^3\text{RE}$ to estimate the MI between two $d$-dimensional correlated Gaussian distributions. 
Specifically, we consider $q(\mathbf{y})=\mathcal{N}(\mathbf{0},\sigma^2\mathbf{E}_d)$ and $p(\mathbf{x})=\mathcal{N}(\mathbf{0},\mathbf{E}_d)$, where $\sigma^2=1e-6$ and $d=\{40, 80, 120 \}$.
Let $p(\mathbf{x},\mathbf{y})$ be the joint density of $\mathbf{X}$ and $\mathbf{Y}$. 
The MI between $\mathbf{X}$ and $\mathbf{Y}$ is defined as $\text{MI}(\mathbf{X},\mathbf{Y})=\mathbb{E}_{\mathbf{x},\mathbf{y}\sim p(\mathbf{x},\mathbf{y})}\left[\log\frac{p(\mathbf{x},\mathbf{y})}{p(\mathbf{x})q(\mathbf{y})}\right]$, and can be approximated via DRE. 
We adapt the experimental setup of \citep{choi2022density} to implement $\text{D}^3\text{RE}$. 

To construct the joint distribution, we use  $q_0(\mathbf{x})=\mathcal{N}(\mathbf{0},\mathbf{E}_d)$ and $q_1(\mathbf{x})=\mathcal{N}(\mathbf{0},\Sigma)$, where $\Sigma$ is block diagonal with $\Lambda=\left[\left[1,\rho\right],\left[\rho,1\right] \right]$  as $2\times 2$ sub-matrices.
For $q_1$, it is designed as a multivariate normal distribution with a block diagonal covariance matrix along the block diagonal. 
Each $\Lambda$ represents the covariance between variable pairs, while off-diagonal blocks remain zero, ensuring no correlation across pairs.
The DDBI and DSBI are implemented, given by  $\mathbf{X}_t^\prime=\alpha_t\mathbf{X}_0+\beta_t\mathbf{X}_1+\sqrt{t(1-t)\gamma^2+(\alpha_t^2+\beta_t^2)\varepsilon}\mathbf{Z}_t$, where $\mathbf{X}_0\sim q_0(\mathbf{x}), \mathbf{X}_1\sim q_1(\mathbf{x})$, and $\mathbf{Z}_t\sim \mathcal{N}(\mathbf{0},\mathbf{E}_d)$.
We estimate the density ratio $r(\mathbf{x})=\frac{q_1(\mathbf{x})}{q_0(\mathbf{x})}$, yielding $\text{MI}(\mathbf{X},\mathbf{Y})\approx\mathbb{E}_{\mathbf{x}\sim q_1(\mathbf{x})}[\log r(\mathbf{x})]$. 

We train the score model using the joint score matching loss (details in \cref{appendix:joint-score-matching}).   
The batch size is set to 512 for $d=\{40,80,160\}$ and 256 for $d=320$, with iteration steps of $\{40k, 100k, 400k, 500k \}$, respectively. 
DRE-$\infty$ serves as the baseline method. 
Results, shown in \cref{fig:MI-evolution}, demonstrate that $\text{D}^3\text{RE}$, especially DSBI, produces MI estimates significantly closer and faster to the ground truth compared to the baseline, highlighting its superiority in accurately capturing mutual dependencies between variables.

\subsection{Density Estimation}
\textbf{Energy-based Modeling on MNIST.}
\label{appendix:likelihood-estimation-MNIST}
We applied the proposed $\text{D}^3\text{RE}$ framework to density estimation on the MNIST dataset, leveraging pre-trained energy-based models (EBMs)  \citep{choi2022density}. 
Let $q_1(\mathbf{x})$ denote the MNIST data distribution and $q_0(\mathbf{x})$ a simple noise distribution with three different settings, as reported in \citep{rhodes2020telescoping}: Gaussian noise, Gaussian copula, and Rational Quadratic Neural Spline Flow (RQ-NSF) \citep{durkan2019neural}. 
We applied an modified version of DDBI of the form $\mathbf{X}_t^\prime=\alpha_t\mathbf{X}_0+\beta_t\text{EBM}(\mathbf{X}_1)+\sqrt{t(1-t)\gamma^2+(\alpha_t^2+\beta_t^2)\varepsilon}\mathbf{Z}_t$, where $\mathbf{X}_0\sim q_0(\mathbf{x}), \mathbf{X}_1\sim q_1(\mathbf{x})$, $\mathbf{Z}_t\sim \mathcal{N}(\mathbf{0},\mathbf{E}_d)$, $\alpha_t=\exp\{-0.25(\beta_{\max}-\beta_{\min})t^2-0.5\beta_{\min}t\}$ and $\beta_t=\sqrt{1-\alpha_t^2}$. 
$\beta_{\min}$ and $\beta_{\max}$ are set to 0.1 and 20, respectively. 
The results are reported in bits-per-dimension (BPD), evaluated as $\text{BPD}=-\frac{1}{d\ln{2}}\mathbb{E}_{\mathbf{x}\sim q_1(\mathbf{x})}\left[ \log q_1(\mathbf{x}) \right]$, where the expectation reflects the log-density of the MNIST dataset. 
Exact BPD computation is infeasible for EBMs; therefore, we estimate it using two annealed MCMC methods: Annealed Importance Sampling (AIS) \citep{neal2001annealed} and Reverse Annealed Importance Sampling Estimator (RAISE) \citep{burda2015accurate}. 
\begin{table}[htbp]
\centering
\caption{Comparison of the estimated log-density on MNIST dataset based on pre-trained energy-based models. The results are reported in BPD. Lower is better. The reported results for NCE and TRE are sourced from  \citep{rhodes2020telescoping}.}

%\vskip 0.15in
\begin{tabular}{cccccc}
\toprule
\textbf{Method} & \textbf{Noise type} & \textbf{Noise} & \textbf{Direct} ($\downarrow$) & \textbf{RAISE} ($\downarrow$) & \textbf{AIS} ($\downarrow$) \\
\midrule
NCE & Gaussian & 2.01 & 1.96 & 1.99 & 2.01 \\
TRE & Gaussian & 2.01 & 1.39 & 1.35 & 1.35 \\
DRE-$\infty$  & Gaussian & 2.01 & 1.33 & 1.33 & 1.33 \\
\midrule
DRE-$\infty$+OTR, ours & Gaussian & 2.01 & 1.313 & 1.31 & 1.31 \\
$\text{D}^3\text{RE}$ (DDBI), ours & Gaussian & 2.01 & 1.297 & 1.30 & 1.29 \\
$\text{D}^3\text{RE}$ (DSBI), ours & Gaussian & 2.01 & \textbf{1.293} & 1.29 & 1.29 \\
\midrule
NCE  & Copula & 1.40 & 1.33 & 1.48 & 1.45 \\
TRE & Copula & 1.40 & 1.24 & 1.23 & 1.22 \\
DRE-$\infty$  & Copula & 1.40 & 1.21 & 1.21 & 1.21 \\
\midrule
DRE-$\infty$+OTR, ours & Copula & 1.40 & 1.204 & 1.19 & 1.18 \\
$\text{D}^3\text{RE}$ (DDBI), ours & Copula & 1.40 & 1.193 & 1.19 & 1.19 \\
$\text{D}^3\text{RE}$ (DSBI), ours & Copula & 1.40 & \textbf{1.170} & 1.19 & 1.18 \\
\midrule
NCE  & RQ-NSF & 1.12 & 1.09 & 1.10 & 1.10 \\
TRE  & RQ-NSF & 1.12 & 1.09 & 1.09 & 1.09 \\
DRE-$\infty$  & RQ-NSF & 1.12 & 1.09 & 1.08 & 1.08 \\
\midrule
DRE-$\infty$+OTR, ours & RQ-NSF & 1.12 & 1.072 & 1.07 & 1.06 \\
$\text{D}^3\text{RE}$ (DDBI), ours & RQ-NSF & 1.12 & 1.072 & 1.06 & 1.06 \\
$\text{D}^3\text{RE}$ (DSBI), ours & RQ-NSF & 1.12 & \textbf{1.066} & 1.06 & 1.06 \\
\bottomrule
\end{tabular}
% \vskip -0.1in
\label{tab:bpd-comparison}
\end{table}

\textbf{2-D Synthetic Datasets.}
\label{appendix:likelihood-estimation-trajectory-2d-methods}
In this section, we present density estimation results on eight synthetic datasets for different methods. From left to right, the epochs are 0, 2000, 4000, 6000, 8000, 10000, 12000, 14000, 16000, 18000 and 20000. Corresponding results are shown in  \cref{fig:likelihood-estimation-trajectory-2d-method-swissroll,fig:likelihood-estimation-trajectory-2d-method-circles,fig:likelihood-estimation-trajectory-2d-method-rings,fig:likelihood-estimation-trajectory-2d-method-moons,fig:likelihood-estimation-trajectory-2d-method-8gaussians,fig:likelihood-estimation-trajectory-2d-method-pinwheel,fig:likelihood-estimation-trajectory-2d-method-2spirals,fig:likelihood-estimation-trajectory-2d-method-checkerboard}. 
D$^3$RE (including DDBI and DSBI) achieved the best performance on all datasets and was able to learn the best results with fewer epochs.

\subsection{Ablation Study on $\gamma^2$}
\label{appendix:ablation-study-gamma}

\textbf{Mutual Information Estimation.}
The ablation study on varying $\gamma^2$ values (\cref{fig:MI-evolution-gamma-main}) reveals distinct convergence behaviors in MI estimation across epochs. For all dimensions ($d = \{40, 80, 120\}$), smaller $\gamma^2$ values ($\leq 0.01$) lead to faster initial convergence toward the DRE-$\infty$ baseline, particularly in lower dimensions ($d = 40$). However, excessively small $\gamma^2 = 0.001$ introduces instability in later epochs, causing slight deviations from the baseline. In contrast, larger $\gamma^2$ values ($\geq 0.1$) show slower initial convergence but stabilize over longer training periods, especially in higher dimensions ($d = 120$). Notably, $\gamma^2 = 0.1$ strikes a balance between convergence speed and stability, consistently aligning with the baseline across all dimensions. These findings suggest that the optimal $\gamma^2$ selection is influenced by both the dimensionality and training duration, with moderate regularization ($\gamma^2 = 0.01$–$0.1$) providing robust MI estimation performance.
% \begin{figure}[htbp]
%     \centering
%     \subfigure[$d=40$]{\includegraphics[width=0.32\linewidth]{gmm_mi_40d_param_joint_gamma.pdf}\label{fig:MI-40d-gamma-appendix}}
%     \hfill
%     \subfigure[$d=80$]{\includegraphics[width=0.32\linewidth]{gmm_mi_80d_param_joint_gamma.pdf}\label{fig:MI-80d-gamma-appendix}}
%     \hfill
%     \subfigure[$d=120$]{\includegraphics[width=0.32\linewidth]{gmm_mi_120d_param_joint_gamma.pdf}\label{fig:MI-120d-gamma-appendix}}
%     \hfill
%     \caption{Evolution of estimated MI across epochs with varying $\gamma^2=\{0.001,0.01,0.1,1.0\}$ and dimensions $d = \{40, 80, 120\}$. The DRE-$\infty$ \citep{choi2022density} is regarded as the `baseline' in this experiment.}
%     \label{fig:MI-evolution-gamma}
% \end{figure}

\textbf{Density Estimation.}
The ablation study on $\gamma^2$ for density estimation (\cref{fig:ablation-study-all-toy}) reveals systematic trade-offs in performance across regularization strengths. For small $\gamma^2 = 0.001$, the model achieves rapid initial alignment with the ground truth distribution (first row) but exhibits overfitting artifacts in later epochs, manifesting as irregular density peaks and deviations from the smooth ground truth structure. Intermediate values ($\gamma^2 = 0.01$–$0.1$) demonstrate balanced behavior: $\gamma^2 = 0.01$ preserves finer details while maintaining stability, and $\gamma^2 = 0.1$ produces smoother approximations with minimal divergence from the true distribution. Larger $\gamma^2$ values ($\geq 0.5$) induce excessive regularization, leading to oversmoothed estimates that fail to capture critical modes of the 2-D data, particularly in high-density regions. Notably, $\gamma^2 = 0.1$ achieves the closest visual and structural resemblance to the ground truth, suggesting its suitability for low-dimensional tasks requiring both fidelity and robustness. These results underscore the necessity of tuning $\gamma^2$ to mitigate under-regularization artifacts while preserving distributional complexity.
% \begin{figure}[ht]
%     \centering
%     \includegraphics[width=0.6\linewidth]{all_likelihood_gamma.pdf}
%     \caption{Ablation study on the effect of $\gamma^2$ for density estimation on 2-D toy data. The first row displays the results for the ground truth data. Each subsequent row, from top to bottom, corresponds to $\gamma^2$ values of 0.001, 0.01, 0.1, 0.5, and 1.0, respectively.}
%     \label{fig:ablation-study-all-toy}
% \end{figure}

We also present density estimation results on eight synthetic datasets for varing values of $\gamma^2$. From left to right, the epochs are 0, 2000, 4000, 6000, 8000, 10000, 12000, 14000, 16000, 18000 and 20000. Corresponding results are shown in  \cref{fig:likelihood-estimation-trajectory-2d-gamma-swissroll,fig:likelihood-estimation-trajectory-2d-gamma-circles,fig:likelihood-estimation-trajectory-2d-gamma-rings,fig:likelihood-estimation-trajectory-2d-gamma-moons,fig:likelihood-estimation-trajectory-2d-gamma-8gaussians,fig:likelihood-estimation-trajectory-2d-gamma-pinwheel,fig:likelihood-estimation-trajectory-2d-gamma-2spirals,fig:likelihood-estimation-trajectory-2d-gamma-checkerboard}.

\subsection{Ablation Study on OTR}
\label{appendix:ablation-study-OTR}

\cref{fig:MI-evolution} compares the MI estimation performance of DDBI and DSBI across different dimensions ($d=80, 120$). DDBI uses diffusion bridges and Gaussian dequantization, while DSBI adds OTR to achieve better alignment. In panel (a), both DDBI and DSBI outperform the baseline methods. DDBI shows stable performance and converges close to the ground truth. However, DSBI, with OTR, achieves faster convergence and higher accuracy, staying closer to the ground truth throughout training. In panel (b), the impact of OTR becomes more evident. Although DDBI still outperforms the baseline methods, its convergence is slower, and its accuracy is lower compared to DSBI. By leveraging OTR, DSBI demonstrates superior MI estimation performance across all training epochs and remains closer to the ground truth in high-dimensional settings ($d=120$).
OTR significantly improves the alignment of intermediate distributions and enhances model performance. When combined with diffusion bridges and Gaussian dequantization, as in DSBI, OTR achieves its full potential. It allows the model to estimate complex distributions more accurately. 

\textbf{Number of Function Evaluations. }
We analyze the impact of OTR on NFE, noting that DI and DDBI do not utilize OTR. 
Our observations show that applying OTR significantly reduces NFE. 
\cref{fig:nfe-comparison-density-ratio-toy2toy} compares NFE across four methods in DRE, highlighting substantial variations in computational efficiency. 
The first approach exhibits the highest NFE, indicating reliance on iterative procedures requiring repeated function evaluations. The second approach achieves a moderate reduction in NFE, likely by minimizing redundant evaluations through minimized transport costs. 

\begin{figure}[ht]
    \centering
    \includegraphics[width=0.8\linewidth]{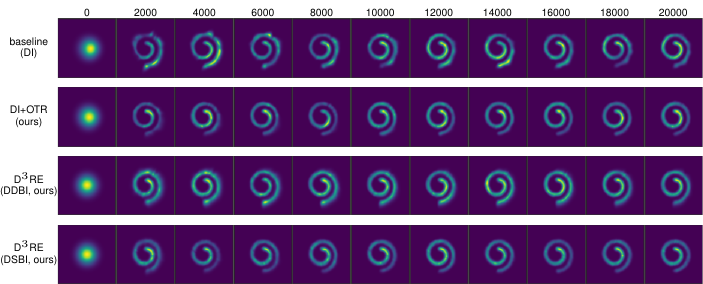}
    \caption{Density estimation results on $\mathsf{swissroll}$ for different methods during training.}
    \label{fig:likelihood-estimation-trajectory-2d-method-swissroll}
\end{figure}
\begin{figure}[ht]
    \centering
    \includegraphics[width=0.8\linewidth]{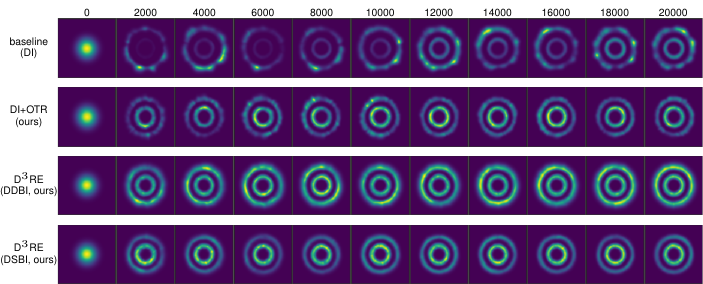}
    \caption{Density estimation results on $\mathsf{circles}$ for different methods during training.}
    \label{fig:likelihood-estimation-trajectory-2d-method-circles}
\end{figure}
\begin{figure}[ht]
    \centering
    \includegraphics[width=0.8\linewidth]{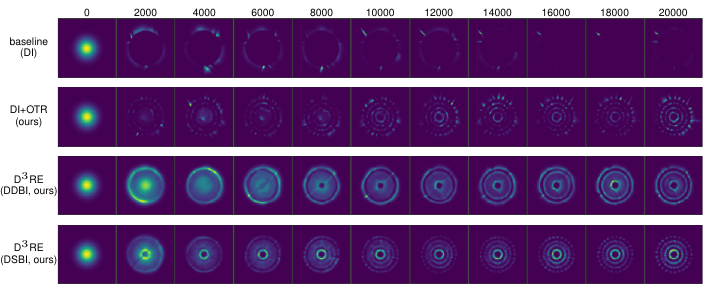}
    \caption{Density estimation results on $\mathsf{rings}$ for different methods during training.}
    \label{fig:likelihood-estimation-trajectory-2d-method-rings}
\end{figure}
\begin{figure}[ht]
    \centering
    \includegraphics[width=0.8\linewidth]{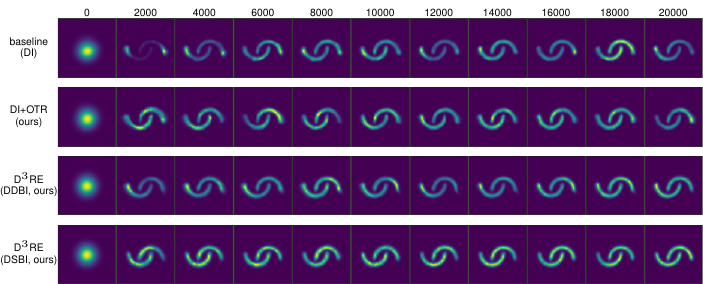}
    \caption{Density estimation results on $\mathsf{moons}$ for different methods during training.}
    \label{fig:likelihood-estimation-trajectory-2d-method-moons}
\end{figure}
\begin{figure}[ht]
    \centering
    \includegraphics[width=0.8\linewidth]{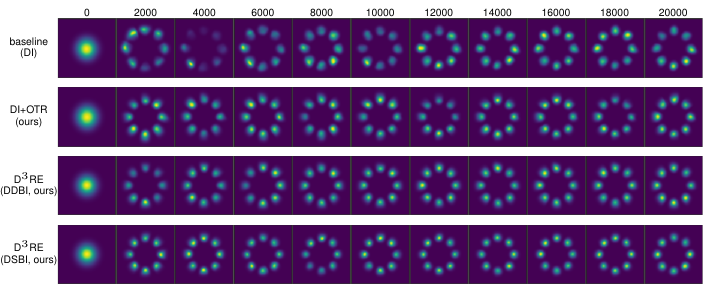}
    \caption{Density estimation results on $\mathsf{8gaussians}$ for different methods during training.}
    \label{fig:likelihood-estimation-trajectory-2d-method-8gaussians}
\end{figure}
\begin{figure}[ht]
    \centering
    \includegraphics[width=0.8\linewidth]{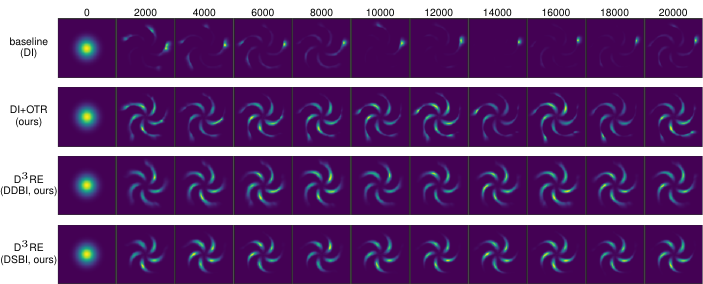}
    \caption{Density estimation results on $\mathsf{pinwheel}$ for different methods during training.}
    \label{fig:likelihood-estimation-trajectory-2d-method-pinwheel}
\end{figure}
\begin{figure}[ht]
    \centering
    \includegraphics[width=0.8\linewidth]{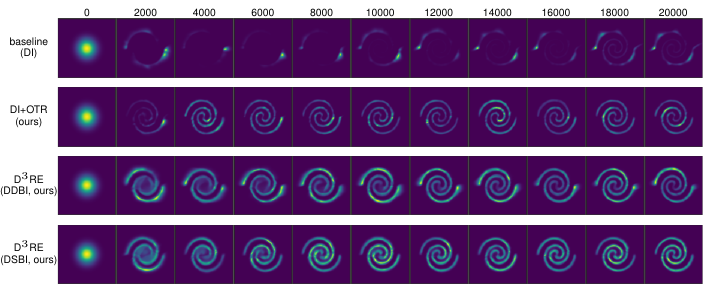}
    \caption{Density estimation results on $\mathsf{2spirals}$ for different methods during training.}
    \label{fig:likelihood-estimation-trajectory-2d-method-2spirals}
\end{figure}
\begin{figure}[ht]
    \centering
    \includegraphics[width=0.8\linewidth]{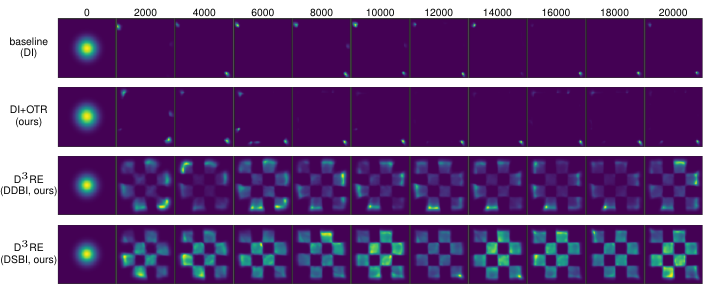}
    \caption{Density estimation results on $\mathsf{checkerboard}$ for different methods during training.}
    \label{fig:likelihood-estimation-trajectory-2d-method-checkerboard}
\end{figure}

\begin{figure}[ht]
    \centering
    \includegraphics[width=0.7\linewidth]{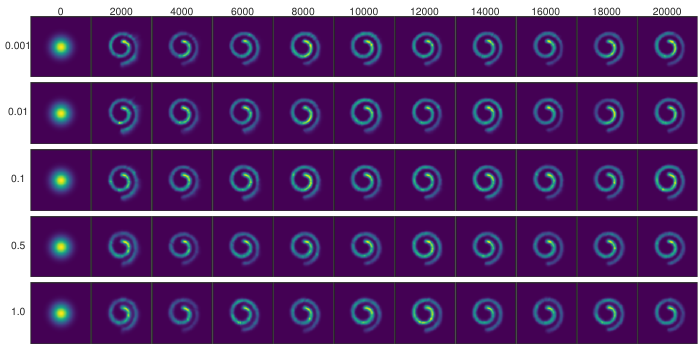}
    \caption{Density estimation results on $\mathsf{swissroll}$ for varing values of $\gamma^2$ during training.}
    \label{fig:likelihood-estimation-trajectory-2d-gamma-swissroll}
\end{figure}

\begin{figure}[ht]
    \centering
    \includegraphics[width=0.70\linewidth]{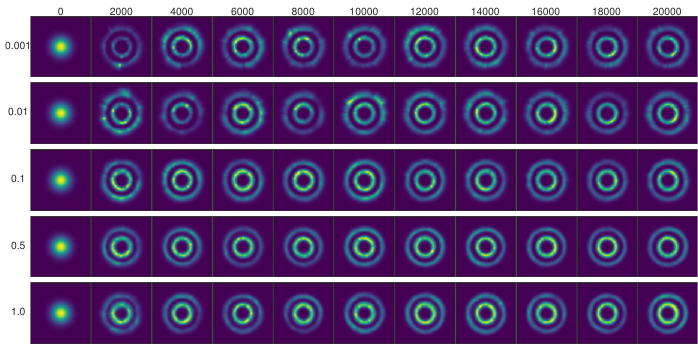}
    \caption{Density estimation results on $\mathsf{circles}$ for varing values of $\gamma^2$ during training.}
    \label{fig:likelihood-estimation-trajectory-2d-gamma-circles}
\end{figure}

\begin{figure}[ht]
    \centering
    \includegraphics[width=0.70\linewidth]{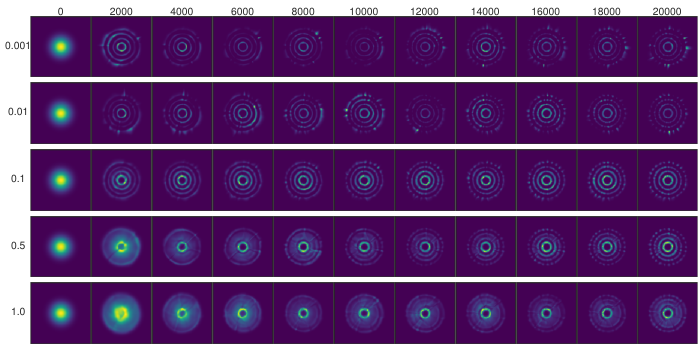}
    \caption{Density estimation results on $\mathsf{rings}$ for varing values of $\gamma^2$ during training.}
    \label{fig:likelihood-estimation-trajectory-2d-gamma-rings}
\end{figure}

\begin{figure}[ht]
    \centering
    \includegraphics[width=0.70\linewidth]{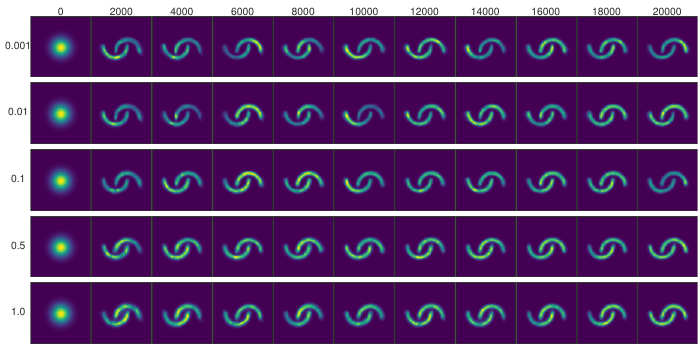}
    \caption{Density estimation results on $\mathsf{moons}$ for varing values of $\gamma^2$ during training.}
    \label{fig:likelihood-estimation-trajectory-2d-gamma-moons}
\end{figure}

\begin{figure}[ht]
    \centering
    \includegraphics[width=0.70\linewidth]{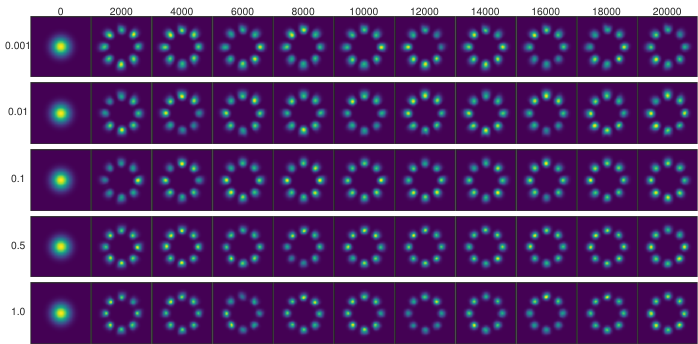}
    \caption{Density estimation results on $\mathsf{8gaussians}$ for varing values of $\gamma^2$ during training.}
    \label{fig:likelihood-estimation-trajectory-2d-gamma-8gaussians}
\end{figure}

\begin{figure}[ht]
    \centering
    \includegraphics[width=0.70\linewidth]{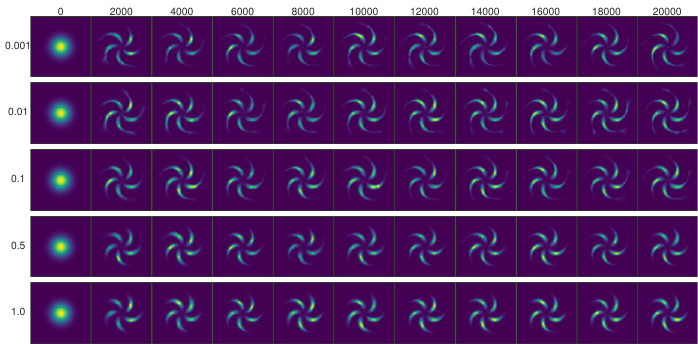}
    \caption{Density estimation results on $\mathsf{pinwheel}$ for varing values of $\gamma^2$ during training.}
    \label{fig:likelihood-estimation-trajectory-2d-gamma-pinwheel}
\end{figure}

\begin{figure}[ht]
    \centering
    \includegraphics[width=0.70\linewidth]{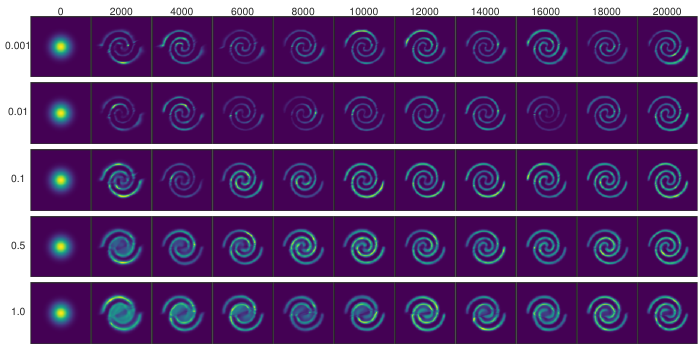}
    \caption{Density estimation results on $\mathsf{2spirals}$ for varing values of $\gamma^2$ during training.}
    \label{fig:likelihood-estimation-trajectory-2d-gamma-2spirals}
\end{figure}

\begin{figure}[ht]
    \centering
    \includegraphics[width=0.70\linewidth]{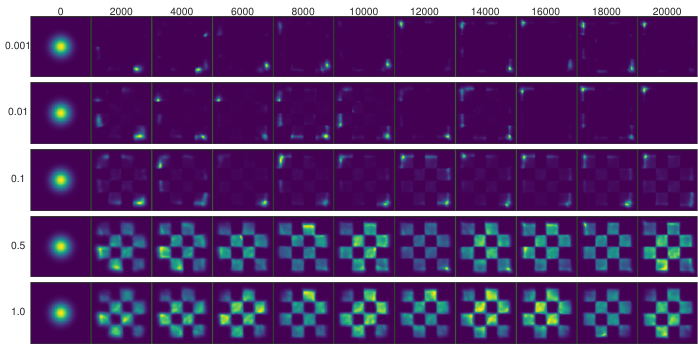}
    \caption{Density estimation results on $\mathsf{checkerboard}$ for varing values of $\gamma^2$ during training.}
    \label{fig:likelihood-estimation-trajectory-2d-gamma-checkerboard}
\end{figure}

%%%%%%%%%%%%%%%%%%%%%%%%%%%%%%%%%%%%%%%%%%%%%%%%%%%%%%%%%%%%%%%%%%%%%%%%%%%%%%%
%%%%%%%%%%%%%%%%%%%%%%%%%%%%%%%%%%%%%%%%%%%%%%%%%%%%%%%%%%%%%%%%%%%%%%%%%%%%%%%

\end{document}